\documentclass{article}
\usepackage{geometry, amssymb, amsmath, graphicx,epstopdf,booktabs}
 \usepackage{fullpage,verbatim}
 \usepackage{pseudocode}
 \usepackage{hyperref} 
 \hypersetup{colorlinks=true,
    linkcolor= blue,          
    citecolor= magenta,        
    filecolor=green,      
    urlcolor=red           
}
\newcommand{\paren}[1]{\left ( #1 \right )}
\newcommand{\brac}[1]{\left [ #1 \right]}
\newcommand{\half}{\frac{1}{2}}

\newcommand{\s}[1]{
  \begin{equation*}
    \begin{split}
      #1
    \end{split}
  \end{equation*}
}
\newcommand{\sn}[2]{
  \begin{equation} \label{#1}
    \begin{split}
      #2
    \end{split}
  \end{equation}
}
\newcommand{\f}[2]{\frac{#1}{#2}}

\newcommand{\giv}{\, \lvert \,}
\newcommand{\Giv}{\, \Bigg\lvert \,}

\newcommand{\dd}{\, \mathrm{d}}

\newcommand{\bhat}{\widehat{\beta}}

\newcommand{\norm}[1]{\left\Vert#1\right\Vert}
\newcommand{\al}{\alpha}
\newcommand{\eps}{\epsilon}
\newcommand{\abs}[1]{\left\vert#1\right\vert}

\newcommand{\ceiling}[1]{\left\lceil#1\right\rceil}
\newcommand{\set}[1]{\left\{#1\right\}}

\newcommand{\model}{\mathcal{M}}
\newcommand{\data}{\mathcal{D}}

\begin{document}

\title{A Minimum Description Length Approach to Multitask Feature Selection}
\author{Brian Tomasik}
\date{May 2009}

\maketitle


\pagestyle{plain}

\begin{abstract}
One of the central problems in statistics and machine learning is \emph{regression}: Given values of input variables, called \emph{features}, develop a model for an output variable, called a \emph{response} or \emph{task}. In many settings, there are potentially thousands of possible features, so that \emph{feature selection} is required to reduce the number of predictors used in the model. Feature selection can be interpreted in two broad ways. First, it can be viewed as a means of \emph{reducing prediction error} on unseen test data by improving model generalization. This is largely the focus within the machine-learning community, where the primary goal is to train a highly accurate system. The second approach to feature selection, often of more interest to scientists, is as a form of \emph{hypothesis testing}: Assuming a ``true" model that generates the data from a small number of features, determine which features actually belong in the model. Here the metrics of interest are precision and recall, more than  test-set error.

Many regression problems involve not one but several response variables. Often the responses are suspected to share a common underlying structure, in which case it may be advantageous to share information across the responses; this is known as \emph{multitask learning}. As a special case, we can use multiple responses to better identify shared predictive features---a project we might call \emph{multitask feature selection}.

This thesis is organized as follows. Section \ref{ch:intro} introduces feature selection for regression, focusing on $\ell_0$ regularization methods and their interpretation within a Minimum Description Length (MDL) framework. Section \ref{regression-section} proposes a novel extension of MDL feature selection to the multitask setting. The approach, called the ``Multiple Inclusion Criterion" (MIC), is designed to borrow information across regression tasks by more easily selecting features that are associated with multiple responses. We show in experiments on synthetic and real biological data sets that MIC can reduce prediction error in settings where features are at least partially shared across responses. Section \ref{hypothesis} surveys hypothesis testing by regression with a single response, focusing on the parallel between the standard Bonferroni correction and an MDL approach. Mirroring the ideas in Section \ref{regression-section}, Section \ref{hypothesis-mic} proposes a novel MIC approach to hypothesis testing with multiple responses and shows that on synthetic data with significant sharing of features across responses, MIC outperforms standard FDR-controlling methods in terms of finding true positives for a given level of false positives. Section \ref{conclusion} concludes.
\end{abstract}

\pagenumbering{arabic}

\section{Feature Selection with a Single Response}\label{ch:intro}

The standard \emph{linear regression model} assumes that a response $y$ is generated as a linear combination of $m$ predictor variables (``features") $x_1$, $\ldots$, $x_{m}$ with some random noise:
\sn{1.1}{
y = \beta_1 x_1 + \beta_2 x_2 + \ldots + \beta_{m} x_{m} + \eps, \ \ \ \eps \sim \mathcal{N}(0, \sigma^2),
}
where we assume the first feature $x_1$ is an intercept whose value is always 1. Given $n$ observations of the features and responses, we can write the $y$ values in an $n \times 1$ vector $Y$ and the $x$ values in an $n \times m$ matrix $X$. Assuming the observations are independent and identically distributed, \eqref{1.1} can be rewritten as
\sn{single-resp-eq}{
Y = X \beta + \eps, \ \ \ \eps \sim \mathcal{N}_n(0, \sigma^2 I_{n \times n}),
}
where $\beta = \begin{bmatrix} \beta_1\\ \ldots\\ \beta_m \end{bmatrix}$, $\mathcal{N}_n$ denotes the $n$-dimensional Gaussian distribution, and $I_{n \times n}$ is the $n \times n$ identity matrix.

The maximum-likelihood estimate $\bhat$ for $\beta$ under this model can be shown to be the one minimizing the residual sum of squares:
\sn{1.3}{
\text{RSS} := (Y - X \beta)' (Y - X \beta).
}
The solution is given by
\sn{1.4}{
\bhat = \paren{X' X}^{-1} X' Y
}
and is called the \emph{ordinary least squares} (OLS) estimate for $\beta$.

In some cases, we may want to restrict the number of features in our model to a subset of $q$ of them, including the intercept. In this case, we pretend that our $X$ matrix contains only the relevant $q$ columns when applying \eqref{1.4}; the remaining $m - q$ entries of the $\bhat$ matrix are set to 0. I'll denote the resulting estimate by $\bhat_q$, and the RSS for that model by
\sn{rssq}{
\text{RSS}_q := (Y - X \bhat_q)' (Y - X \bhat_q).
}

\subsection{Penalized Regression}\label{Penalized Regression}

In many cases, regression problems have large numbers of potential features. For instance, \cite{foster2004vsd} predicted credit-card bankruptcy using a model with more than $m = 67{,}000$ potential features. In bioinformatics applications, it is common to have thousands or tens of thousands of features for, say, the type of each of a number of genetic markers or the expression levels of each of a number of gene transcripts. The number of observations, in contrast, is typically a few hundred at best.

The OLS esimate \eqref{1.4} breaks down when $m > n$, since the $m \times m$ matrix $X' X$ is not invertible due to having rank at most $n$. Moreover, it's implausible that a given response is \textit{actually} linearly related to such a large number of features; a model $\bhat$ with lots of nonzeros is probably overfitting the training data.

The statistics and machine-learning communities have developed a number of approaches for addressing this problem. One of the most common is \emph{regularized regression}, which aims to minimize not \eqref{1.3} directly, but a penalized version of the residual sum of squares:
\sn{regularized}{
(Y - X \beta)' (Y - X \beta) + \lambda \norm{\beta}_p,
}
where $\norm{\beta}_p$ represents the $\ell_p$ norm of $\beta$ and $\lambda$ is a tunable hyperparameter, to be determined by cross-validation or more sophisticated \emph{regularization path} approaches.\footnote{See, e.g., \cite{friedman2008rpg} for an excellent introduction.}

\emph{Ridge regression} takes the penalty as proportional to the (square of the) $\ell_2$ norm: $\lambda \norm{\beta}_2^2$. This corresponds to a Bayesian maximum a posteriori estimate for $\beta$ under a Gaussian prior $\beta \sim \mathcal{N}(0_{m \times 1}, \f{\sigma^2}{\lambda} I_{m \times m})$, with $\sigma^2$ as in \eqref{1.1} \cite[p. 153]{Bishop:Pattern}. Under this formulation, we have
\s{
\bhat = (X' X + \lambda I_{m \times m})^{-1} X' Y,
}
which is computationally valid because $X ' X + \lambda I_{m \times m}$ is invertible for $\lambda > 0$. However, because the square of a decimal number less than one is much smaller than the original number, the $\ell_2$ norm offers little incentive to drive entries of $\bhat$ to 0---many of them just become very small.

Another option is to penalize by the $\ell_1$ norm, which is known as \emph{lasso regression} and is equivalent to a double-exponential prior on $\beta$ \cite{tibshirani1996rsa}. Unlike $\ell_2$, $\ell_1$ regularization doesn't square the coefficients, and hence the entries of $\bhat$ tend to be sparse. In this way, $\ell_1$ regression can be seen as a form of feature selection---i.e., choosing a subset of the original features to keep in the model (e.g., \cite{blum1997srf}). Sparsity helps to avoid overfitting to the training set; as a result, the number of training examples required for successful learning with $\ell_1$ regularization grows only logarithmically with the number of irrelevant features, whereas this number grows linearly for $\ell_2$ regression \cite{ng2004fsv}. Sparse models also have the benefit of being more interpretable, which is important for scientists who want to know which particular variables are actually relevant for a given response. Building regression models for interpretation is further discussed in Sections \ref{hypothesis} and \ref{hypothesis-mic}.

If $\ell_2$ regression fails to achieve sparsity because coefficients are squared, then, say, $\ell_\half$ regression should achieve even more sparsity than $\ell_1$. As $p$ approaches 0, $\norm{\beta}_p$ approaches the number of nonzero values in $\beta$. Hence, regularization with what is called the ``$\ell_0$ norm" is \emph{subset selection}: Choosing a small number of the original features to retain in the regression model. Once a coefficient is in the model, there's no incentive to drive it to a small value; all that counts is the cost of adding it in the first place. The $\ell_0$ norm has a number of advantages \cite{defensel0}, including bounded worst-case risk with respect to the $\ell_1$ norm and better control of a measure called the ``false discover rate" (FDR), explained more fully in Section \ref{hypothesis}. 
Moreover, as \cite[p. 1]{obozinski} note, ``A virtue of the [$\ell_0$] approach is that it focuses on the qualitative decision as to whether a covariate is relevant to the problem at hand, a decision which is conceptually distinct from parameter estimation." However, they add, ``A virtue of the [$\ell_1$] approach is its computational tractability." Indeed, exact $\ell_0$ regularization requires subset search, which has been proved NP-hard \cite{natarajan1995sas}. In practice, therefore, an approximate greedy algorithm like forward stepwise selection is necessary \cite[p. 582]{chapter-maimon2005dma}.

In a regression model, residual sum of squares is proportional up to an additive constant to the negative log-likelihood of $\beta$. Therefore, $\ell_0$ regularization can be rephrased as a \emph{penalized likelihood} criterion (with a different $\lambda$ than in \eqref{regularized}):
\sn{penlike}{
-2 \ln P(Y \giv \bhat_q) + \lambda q,
}
where $q$ is the number of features in the model, and $P(Y \giv \bhat_q)$ is the likelihood of the data given a model containing $q$ features. As noted in \cite{george2000cae}, statisticians have proposed a number of choices for $\lambda$, including
\begin{itemize}
\item $\lambda = 2$, corresponding to Mallows' $C_p$ \cite{mallows1973scc}, or approximately the Akaike Information Criterion (AIC) \cite{akaike73},
\item $\lambda = \ln n$, the Bayesian Information Criterion (BIC) \cite{schwartz79}, and
\item $\lambda = 2 \ln m$, the Risk Inflation Criterion (RIC) \cite{donohojohnstone94, fostergeorge94}.
\end{itemize}
It turns out that each of these criteria can be derived from an information-theoretic principle called Minimum Description Length (MDL) under different ``model coding" schemes (e.g., \cite{foster1999lac, hansen1999baa}). MDL forms the focus of this thesis, and its approach to regression is the subject of the next subsection.

\subsection{Regression by Minimum Description Length}\label{singleResponse}

MDL \cite{rissanen1978, rissanen1999} 
is a method of model selection that treats the best model as the one which maximally compresses a digital representation of the observed data. We can imagine a ``Sender" who wants to transmit some data in an email to a ``Receiver" using as few bits as possible \cite{stine2004}. In the case of linear regression, we assume that both Sender and Receiver know the $n \times m$ matrix $X$, and Sender wants to convey the values in the $n \times 1$ matrix $Y$. 
One way to do this would be to send the raw values for each of the $n$ observations $Y = \begin{bmatrix} y_1 \\ \ldots \\ y_n \\ \end{bmatrix}$. However, if the response is correlated with some features, it may be more efficient first to describe a regression model $\widehat{\beta}$ for $Y$ given $X$ and then to enumerate the residuals $Y - X \widehat{\beta}$, which---having a narrower distribution---require fewer bits to encode.

To minimize description length, then, Sender should choose $\widehat{\beta}$ to minimize
\sn{dl}{
\mathcal{D}(Y \giv \widehat{\beta}) + \mathcal{D}(\widehat{\beta}),
}
where the first term is the description  length of the residuals about the model, and the second term is the description length of the model itself.\footnote{This is what \cite[p. 11]{grunwald2005} calls the ``crude, two-part version of MDL." Beginning with \cite{rissanen1986}, Rissanen introduced a ``one-part" version of MDL based on a concept called ``stochastic complexity." However, it too divides the description length into terms for model fit and model complexity, so it is in practice similar to two-part MDL \cite{stine2004}. } Exactly what these terms mean will be elaborated below.

\subsubsection{Coding the Data: $\mathcal{D}(Y \giv \widehat{\beta})$}\label{Coding the Data}

The Kraft inequality in information theory \cite[sec. 5.2]{cover2006} implies that for any probability distribution $\set{p_i}$ over a finite or countable set, there exists a corresponding code with codeword lengths $\ceiling{-\lg p_i}$. Moreover, these code lengths are optimal in the sense of minimizing the expected code length with respect to $\set{p_i}$. If Sender and Receiver agree on a model for the data, e.g., \eqref{1.1}, then they have a probability distribution over possible configurations of residuals $\eps$, so they will agree to use a code for the residuals with lengths
\sn{yGivenBeta}{
\mathcal{D}(Y \giv \widehat{\beta}) = - \lg P(\eps \giv \bhat) = - \lg P(Y \giv \bhat),
}
that is, the negative log-likelihood of the data given the model. This is a standard statistical measure for poorness of fit.\footnote{We use ``idealized" code lengths and so drop the ceiling on $- \lg P(Y \giv \bhat)$ \cite[p. 2746]{barron1998}
. Also, since $P(Y \giv \bhat)$ is a continuous density, we would in practice need to discretize it. This could be done to a given precision with only a constant impact on code length \cite[sec. 8.3]{cover2006}.}

Consider a stepwise-regression scenario in which our model currently contains $q-1$ features (including the intercept term), and we're deciding whether to include an extra $q^\text{th}$ feature. Let $Y_i$ denote the $i^\text{th}$ entry (row) of $Y$ and $\bhat_q$ a model with all $q$ features. \eqref{single-resp-eq} and \eqref{yGivenBeta} give
\sn{lengthIfKnowSigma}{
\mathcal{D}(Y \giv \widehat{\beta}_q) &= - \lg \prod_{i=1}^n P(Y_i \giv \bhat_q)\\
&= - \sum_{i=1}^n \lg \brac{ \f{1}{\sqrt{2 \pi \sigma^2}} \exp\paren{-\f{1}{2 \sigma^2} (Y_i - X \bhat_q)^2 }}\\
&= \f{1}{2 \ln 2} \brac{ n \ln (2 \pi \sigma^2) + \f{\text{RSS}_q}{\sigma^2} },
}
where $\text{RSS}_q$ is as in \eqref{rssq}. Of course, $\sigma^2$ is in practice unknown, so we estimate it by\footnote{
This is the maximum-likelihood estimate for $\sigma^2$, which Sender uses because, ignoring model-coding cost, maximizing likelihood is equivalent to minimizing description length. In practice, of course, many statisticians use the unbiased estimate 
\s{
\widehat{\sigma}^2 = \f{\text{RSS}_q}{n-q}.
}
\cite[Appendix A]{stine2004} makes an interesting case for this latter value within the MDL framework. However, we use the maximum-likelihood estimate for its elegance and convenience.}
\sn{sigmaHatEst}{
\widehat{\sigma}^2 = \f{\text{RSS}_q}{n}.
}
Inserting this estimate for $\sigma^2$ in \eqref{lengthIfKnowSigma} gives
\sn{finalLogLike}{
\mathcal{D}(Y \giv \widehat{\beta}_q) &= \frac{1}{2 \ln 2} \brac{ n \ln (2 \pi \widehat{\sigma}^2) + n}\\
&= \f{n}{2 \ln 2} \brac{  \ln \paren{ \f{2 \pi  \text{RSS}_q}{n}} + 1}.
}

Unfortunately, in practice, \eqref{sigmaHatEst} seems to cause overfitting, especially when used for multiple responses as in Section \ref{logLikeMultipleY}. The overfitting comes from the fact that $\widehat{\sigma}^2$ assumes that the current $q^\text{th}$ feature actually comes into the model; if this is not the case, the estimated variance is spuriously too low, which allows random noise to appear more unlikely than it really is. To prevent overfitting, we use instead an estimate of $\widehat{\sigma}^2$ based on the model without the current feature: $\bhat_{q-1}$. That is,
\sn{sigmaHatEstNull}{
\widehat{\sigma}^2 = \f{1}{n} \sum_{i=1}^n (Y_i - X \bhat_{q-1})^2 = \f{\text{RSS}_{q-1}}{n} ,
}
which means that \eqref{lengthIfKnowSigma} becomes instead
\sn{finalLogLikeNull}{
\mathcal{D}(Y \giv \widehat{\beta}_q) &= \f{n}{2 \ln 2} \brac{  \ln \paren{ \f{2 \pi \text{RSS}_{q-1} }{n}} + \f{\text{RSS}_q}{\text{RSS}_{q-1}} }.\\
}

\subsubsection{Coding the Model: $\mathcal{D}(\widehat{\beta})$}\label{codeModel}

Just as $\mathcal{D}(Y \giv \widehat{\beta})$ depends on the model for the residuals that Sender and Receiver choose, so their coding scheme for $\bhat$ itself will reflect their prior expectations.\footnote{Again by the Kraft inequality, we can interpret $2^{-\mathcal{D}(\beta)}$ as a prior over possible models $\beta$. In fact, this is done explicitly in the Minimum Message Length (MML) principle (e.g., \cite{wallace2005sai}), a Bayesian sister to MDL which chooses the model $\bhat$ with maximum $P(\beta \giv Y)$, i.e., the model that minimizes
\s{
-\lg P(\beta \giv Y) = -\lg P(\beta) - \lg P(Y \giv \beta) + \text{const}.
}
}
When the number of features $m$ is large (say, $1000$), Sender will likely only want to transmit a few of them that are most relevant, and hence the $\bhat$ matrix will contain mostly zeros.\footnote{For convenience, we assume that Sender always transmits the intercept coefficient $\bhat_1$ and that, in fact, doing so is free of charge. Thus, even if Sender were to encode all other feature coefficients as 0, Receiver would at least have the average $\overline{Y}$ of the $Y$ values to use in reconstructing $Y$.} So the first step in coding $\bhat$ could be to say where the nonzero entries are located; if only a few features enter the model, this can be done relatively efficiently by listing the indices of the features in the set $\set{1, 2, \ldots, m}$. This requires $\ceiling{\lg m}$ bits, which we idealize as just $\lg m$.

The second step is to encode the numerical values of those coefficients. Rissanen \cite{rissanen1983} suggested the basic approach for doing this: Create a discrete grid over some possible parameter values, and use a code for integers to specify which grid element is closest.\footnote{Thus, Sender transmits only rounded coefficient values. Rounding adds to the residual coding cost because Sender is not specifying the exact MLE, but the increase tends to be less than a bit \cite{rissanen1989scs}.} \cite{stine2004} described a simple way to approximate the value of a particular coefficient $\bhat$: Encode an integer version of its z-score relative to the null-hypothesis value $\beta_0$ (which in our case is 0):
\sn{codecoeffcost}{
\left\langle \f{\bhat - \beta_0}{\text{SE}(\bhat)} \right\rangle = \left\langle \f{\bhat}{\text{SE}(\bhat)} \right\rangle,
}
where $\langle x \rangle$ means ``the closest integer to $x$." The z-score can then be coded with the idealized universal code for the positive integers of \cite{elias1975} and \cite{rissanen1983}, in which the cost to code $i \in \set{1, 2, 3, \ldots}$ is
\s{
\lg^* i + b,
}
where $\lg^* i := \lg i + \lg \lg i + \lg \lg \lg i + \ldots$ so long as the terms remain positive, and $b \approx \lg 2.865 \approx 1.516$ \cite[p. 424]{rissanen1983} is the constant such that
\s{
\sum_{i=1}^\infty 2^{-(\lg^*i + b)} = 1.
}
We require the $\lg^*$ instead of a simple $\lg$ because the number of bits Sender uses to convey the integer $i$ will vary, and she needs to tell Receiver how many bits to expect. This number of bits is itself an integer that can be coded, hence the iteration of logarithms. The middle row of Table \ref{sampleLogStar} shows example costs with this code.
\begin{table}[h]
\centering
\begin{tabular}{cccccccc}
\toprule 
$i$ & 1 & 2 & 3 & 4 & 5 & 10 & 100\\
\midrule
Universal-code cost for $i$ & 1.5 & 2.5 & 3.8 & 4.5 & 5.3 & 7.4 & 12.9\\
Universal-style cost for $i$, truncated at 1000 & 1.2 & 2.2 & 3.4 & 4.2 & 5.0 & 7.0 & 12.6\\
\bottomrule
\end{tabular}
\caption{Example costs of the universal code for the integers.}
\label{sampleLogStar}
\end{table}

In fact, in practice it's unnecessary to allow our integer code to extend to arbitrarily large integers. We're interested in features near the limit of detectability, and we expect our z-scores to be roughly in the range $\sim 2$ to $\sim 4$, since if they were much higher, the true features would be obvious and wouldn't require sensitive feature selection. We could thus impose some maximum possible z-score $Z$ that we might ever want to encode (say, 1000) and assume that all of our z-scores will fall below it. In this case, the constant $c$ can be reduced to a new value $c_Z$, now only being large enough that 
\sn{definingcz}{
\sum_{i=1}^{Z} 2^{-(\lg^*i + c_{Z})} = 1.
}
In particular, $c_{1000} \approx 1.199$, which is reflected in the reduced costs shown in the third row of Table \ref{sampleLogStar}. As it turns out, $c_Z \approx 1$ over a broad range (say, for $Z \in \set{5, \ldots, 1000}$).

In our implementation, we avoid computing the actual values of our z-scores (though this could be done in principle), instead assuming a constant cost of 2 bits per coefficient. Though MDL theoretically has no tunable parameters once Sender and Receiver have decided on their models, the number of bits to specify a coefficient can act as one, having much the same effect as the significance level $\alpha$ of a hypothesis test (see Section \ref{mdl-corrections} for details). We found 2 bits per coefficient to work well in practice.

\subsubsection{Summary}\label{Summary}

Combining the cost of the residuals with the cost of the model gives the following formula for the description length as a function of the number of features $q$ that we include in the model:
\sn{summary-1-response}{
- \lg P(Y \giv \bhat_q) + q (\lg m + 2).
}
Note the similarity between \eqref{summary-1-response} and the RIC penalty for \eqref{penlike}.

\section{MIC for Prediction}\label{regression-section}

Standard multivariate linear regression as described in Section \ref{ch:intro} assumes many features but only a single response. However, a number of practical problems involve both multiple features and multiple responses. For instance, a biologist may have transcript abundances for thousands of genes that she would like to predict using genetic markers \cite{kendziorski_statistical_2006}, or she may have multiple diagnostic categories of cancers whose presence or absence she would like to predict using transcript abundances \cite{khan2001cad}. \cite{breiman1997pmr} describe a situation in which they were asked to use over 100 econometric variables to predict stock prices in 60 different industries.

More generally, we suppose we have $h$ response variables $y_1, \ldots, y_h$ to be predicted by the same set of features $x_2, \ldots, x_p$ (with $x_1$ being an intercept, as before). In parallel to \eqref{single-resp-eq}, we can write
\sn{modelsenderreceiver}{
Y = X \beta + \eps,
}
where now $Y$ is an $n \times h$ matrix with each response along a column. $X$ remains $n \times m$, while $\beta$ is $m \times h$ and $\eps$ is $n \times h$.

In general, the noise on the responses may be correlated; for instance, if our responses consist of temperature measurements at various locations, taken with the same thermometer, then if our instrument drifted too high at one location, it will have been too high at the other. A second, practical reason we might make this assumption comes from our use of stepwise regression to choose our model: In the early iterations of a stepwise algorithm, the effects of features not present in the model show up as part of the ``noise" error term, and if two responses share a feature not yet in the model, the portion of the error term due to that feature will be the same. Thus, we may decide to take the rows of $\eps$ have nonzero covariance:
\sn{generalsigma}{
\eps_i \sim \mathcal{N}_h(0, \Sigma),
}
where $\eps_i$ is the $i$th row of $\eps$, and $\Sigma$ is an arbitrary (not necessarily diagonal) $h \times h$ covariance matrix. In practice, however, we end up using a diagonal estimate of $\Sigma$, as discussed in Section \ref{logLikeMultipleY}.

\subsection{Multitask Learning}\label{Multitask Learning}

The na\"{i}ve approach to regression with the above model is to treat each response as a separate regression problem and calculate each column of $\bhat$ using just the corresponding column of $Y$. This is completely valid mathematically; it even works in the case where the covariance matrix $\Sigma$ is non-diagonal, because the maximum-likelihood solution for the mean of a multivariate Gaussian is independent of the covariance \cite[p. 147]{Bishop:Pattern}. However, intuitively, if the regression problems are related, we ought to be able to do better by ``borrowing strength" across the multiple responses. This concept goes by various names; in the machine-learning community, it's often known as ``multitask learning" (e.g., \cite{caruana1997ml}) or ``transfer learning" (e.g., \cite{raina2006cip}). 

Several familiar machine-learning algorithms naturally do multitask learning; for instance, neural networks can share hidden-layer weights with multiple responses. In addition, a number of explicit multitask regression methods have been proposed, such as \emph{curds and whey}, which takes advantage of correlation among responses \cite{breiman1997pmr}, and various dimensionality-reduction algorithms (e.g., \cite{abraham2005dra}).

Some techniques even do explicit feature selection. For instance, \cite{brown1998mbv, brown2002bma} present Bayesian Markov Chain Monte Carlo approaches for large numbers of features. \cite{simila2005msr} present a \emph{Multiresponse Sparse Regression} (MRSR) algorithm that selects features in a stepwise manner by choosing at each iteration the feature most correlated with the residuals of the responses from the previous model.\footnote{Another sparse-regression algorithm is BBLasso, described in Section \ref{BBLasso} below.} We have not explored the above algorithms in detail, but comparing them against MIC could be a fruitful direction for future work. Below we elaborate on two multitask-learning approaches that we have used.

\subsubsection{AndoZhang}\label{AndoZhang}

Ando and Zhang \cite{ando2005flp} aim to use multiple prediction problems to learn an underlying shared structural parameter on the input space. Their presentation is general, but they give the particular example of a linear model, which we'll rewrite using the notation of this thesis. Given $h$ tasks and $m$ features per task, the authors assume the existence of an $H \times m$ matrix\footnote{The authors use $h$ instead of $H$, but we have already used that variable to denote the number of tasks.} $\Theta$, with $H < m$, such that for each task $k = 1, \ldots, h$,
\s{
Y^k = X \beta^k + X \Theta' \beta_\Theta^k,
}
where superscript $k$ denotes the $k^\text{th}$ column and $\beta_\Theta$ is an $H \times h$ matrix of weights for the ``lower dimensional features" created by the product $X \Theta'$ (p. 1824). The transformation $\Theta$ is common across tasks and is thus the way we share information. The authors develop an algorithm, referred to as \emph{AndoZhang} in this thesis, that chooses (p. 1825)
\s{
\brac{ \set{ \bhat, \bhat_\Theta }, \widehat{\Theta} } = \underset{\set{\beta, \beta_\Theta}, \Theta}{\operatorname{argmin}}  \sum_{k=1}^h \paren{ \frac{1}{n} \norm{ Y^k - X \paren{\beta^k + \Theta' \beta_\Theta^k} }^2 + \lambda_k \norm{\beta^k}^2 }  \text{ \ such that \ } \Theta \Theta' = I_{H \times H}.
}
Here $\set{\lambda_k}_{k=1}^h$ are regularization weights, and (unlike Ando and Zhang) we've assumed a constant number $n$ of observations for each task, as well as a quadratic loss function. $\norm{ \cdot }^2$ denotes the square of the $\ell_2$ norm.

Ando and Zhang present their algorithm as a tool for semi-supervised learning: Given one particular problem to be solved, it's possible to generate auxiliary problems from the known data, solve them, and then use the shared structural parameter $\Theta$ on the original problem. Of course, the approach works just as well when we start out with all of the tasks as problems to be solved in their own right.

\subsubsection{BBLasso}\label{BBLasso}

A few multitask learning methods actually do feature selection, i.e., building models that contain only a subset of the original features. AndoZhang, with its $\ell_2$ regularization penalty, leaves in most or all of the features, but $\ell_1$ methods, such as \emph{BBLasso} presented below, do not.

 \cite{argyriou} present an algorithm for learning \textit{functions} of input features across tasks. This is more general than the approach considered in this thesis of choosing merely a subset of the original features, but \cite[sec. 2.2]{argyriou} note that their approach reduces to feature selection with an identity function. Regularizing with an $\ell_1$ norm over features, their problem thus becomes
\sn{objective}{
\bhat = \underset{\beta}{\operatorname{argmin}} \paren{ \text{RSS}_\beta + \lambda \sum_{i=1}^m \norm{\beta_i}},
}
where $\text{RSS}_\beta = \text{trace} \brac{ (Y - X \beta)' (Y - X \beta) }$ is the residual sum of squares for all $h$ responses, and $\beta_i$ denotes the $i^\text{th}$ row of $\beta$. The $i^\text{th}$ term in the sum represents the magnitude of coefficients assigned to feature $i$; the sum over these values for each $i$ amounts to an $\ell_1$ norm over the rows.

The $\ell_2$ norm favors shared coefficients across responses for the same features. To see this, \cite[p. 6]{argyriou} suggest the case in which the entries of $\beta$ can only be 0 or 1. If $h$ different features each have a single nonzero response coefficient, the penalty will be $\lambda \sum_{j=1}^h 1 = \lambda h$, since $\norm{\beta_i} = 1$ for each such feature $i$. However, if a single feature shares a coefficient of 1 across all $h$ responses, the penalty is only $\lambda \sqrt{\sum_{j=1}^h 1^2} = \lambda \sqrt{h}$. The same number and magnitude of nonzero coefficients thus leads to a much smaller penalty when the coefficients are shared.

In principle the penalty term can be made even smaller by including nonzero coefficients for only some of the responses. For instance, in the example above, the penalty would be $\sqrt{h/2}$ if just half of the responses had nonzero coefficients (i.e., coefficients of 1). However, in reality, the coefficient values are not restricted to 0 and 1, meaning that unhelpful coefficients, rather than being set to 0, tend to get very small values whose square is negligible in the $\ell_2$-norm penalty. In practice, then, this algorithm adds entire rows of nonzero coefficients to $\bhat$.

Of course, as \cite[p. 4]{obozinski} note, the $\ell_2$ norm used here could be generalized to an $\ell_p$ norm for any $p \in [1, \infty)$. $p=1$ corresponds to ``no sharing" across the responses, since \eqref{objective} would then reduce to the objective function for $h$ independent $\ell_1$-penalized regressions. At the opposite extreme, $p=\infty$ corresponds to ``maximal sharing"; in that case, only the maximum absolute coefficient value across a row matters.

A number of efficient optimization approaches have been proposed for \eqref{objective} with $p=2$ and $p=\infty$; see, e.g., the related-work sections of \cite{turlach2005svs} and \cite{simila2007isa} for a survey.
In particular, \cite{obozinski} propose a fast approximate algorithm using a regularization path, that is, a method which implicitly evaluates \eqref{objective} at all values of $\lambda$.

\subsection{Regression by MIC}\label{Information-Theoretic Regression---Many Responses}

Following Section \ref{singleResponse}, this section considers ways  we might approach regression with multiple responses from the perspective of information theory. One strategy, of course, would be to apply the penalized-likelihood criterion \eqref{summary-1-response} to each response in isolation; we'll call this the \emph{RIC} method, because the dominant penalty in \eqref{summary-1-response} is $\lg m$, the same value prescribed by the Risk Inflation Criterion for equation \eqref{penlike}.

However, following the intuition of BBLasso and other multitask algorithms, we suspect that features predictive of one task are more likely predictive of other, related tasks. For example, if two of our responses are ``level of HDL cholesterol" and ``level of LDL cholesterol," we expect that lifestyle variables correlated with one will tend to be correlated with the other. The following multitask coding scheme, called the \emph{Multiple Inclusion Criterion} (MIC), is designed to take advantage of such situations by making it easier to add to our models features that are shared across responses.

\subsubsection{Coding the Model}\label{multipleYModel}

The way MIC borrows strength across responses is to efficiently specify the feature-response pairs in the $m \times h$ matrix $\bhat$. The na\"{i}ve approach would be to put each of the $m h$ coefficients in a linear order and specify the index of the desired coefficient using $\lg (mh)$ bits. But we can do better. If we expect nearly all the responses to be correlated with the predictive features, we could give all of the responses nonzero coefficients (using $2h$ bits to code the values for each of the $h$ response coefficients) and simply specify the feature that we're talking about using $\lg m$ bits, as in Section \ref{codeModel}. We'll call this the \emph{fully dependent MIC} coding scheme (or \emph{Full MIC} for short). It amounts to adding entire rows of nonzero coefficients at a time to our $\bhat$ matrix, in much the same way that BBLasso does.

In many cases, however, the assumption that each feature is correlated with almost all of the responses is unrealistic. A more flexible coding scheme would allow us to specify only a subset of the responses to which we want to give nonzero coefficients. For instance, suppose we're considering feature number 703; of the $h=20$ responses, we think that only responses $\set{2, 5, 11, 19}$ should have nonzero coefficients with the current feature. We can use $\lg m$ bits to specify our feature (number 703) once, and then we can list the particular responses that have nonzero coefficients with feature 703, thereby avoiding four different $\lg (mh)$ penalties to specify each coefficient in isolation.

A standard way to code a subset of a set of size $h$ is to first specify how many $k \leq h$ elements the subset contains and then which of the $\binom{h}{k}$ possible subsets with $k$ elements we're referring to \cite[sec. 7.2]{cover2006}. In particular, we choose to code $k$ using
\sn{coding-k}{
\lg^* k + c_h
}
bits, with $c_h$ as defined in \eqref{definingcz}.\footnote{We could also assume a uniform distribution on $\set{1, 2, \ldots, h}$ and spend $\lg h$ bits to code $k$'s index. However, in practice we find that smaller values of $k$ occur more often, so that the $\lg^*$-based code is generally cheaper.} We then need $\lg \binom{h}{k}$ additional bits to specify the particular subset. We refer to this code as \emph{partially dependent MIC} (or \emph{Partial MIC} for short).

\subsubsection{Coding the Data}\label{logLikeMultipleY}

As usual, $\mathcal{D}(Y \giv \widehat{\beta}_q) = - \lg P(Y \giv \bhat_q)$ for the model that Sender and Receiver choose, which in this case is \eqref{modelsenderreceiver}. If we allow rows of $\eps$ to have arbitrary covariance as in \eqref{generalsigma}, then
\sn{fullcoveq}{
\mathcal{D}(Y \giv \widehat{\beta}_q) = \f{1}{2 \ln 2} \brac{ n \ln{ \left( (2\pi)^h |\Sigma| \right) } + \sum_{i=1}^n \paren{ Y_i - X_i \bhat_q }'  \Sigma^{-1} \paren{ Y_i - X_i \bhat_q } },
}
with subscript $i$ denoting the $i^{\text{th}}$ row. Since $\Sigma$ is in fact unknown, we estimate it using maximum likelihood:
\sn{sigmaHatF}{
\widehat{\Sigma}_F  = \f{1}{n} \paren{ Y - X \bhat_{q-1} }' \paren{ Y - X \bhat_{q-1} },
}
where the subscript $F$ stands for ``full covariance," and we use $\bhat_{q-1}$ instead of $\bhat_q$ to prevent overfitting, as in the single-response case of Section \ref{Coding the Data}.

In practice, we find that estimating all $h^2$ entries of $\Sigma$ leads to overfitting. Therefore, in our experiments, we estimate $\Sigma$ using a diagonal matrix $\widehat{\Sigma}_D$ that's the same as $\widehat{\Sigma}_F$ except that the off-diagonal entries have been set to 0. In this case, \eqref{fullcoveq} reduces to a sum of $h$ terms like \eqref{finalLogLikeNull} for each response separately. We also experimented with shrunken estimates of the form $\widehat{\Sigma}_\lambda = \lambda \widehat{\Sigma}_D + (1-\lambda) \widehat{\Sigma}_F$ for $\lambda \in [0,1]$ and observed informally  that for $\lambda > 0.5$ or so, $\widehat{\Sigma}_\lambda$ performed comparably with $\widehat{\Sigma}_D$. However, there did not appear to be a performance advantage, so we continued to use $\widehat{\Sigma}_D$, which is faster computationally.\footnote{We also experimented with two other regularized full covariance-matrix estimators, proposed in \cite{ledoit2004wce} and \cite{SchaferEtAl}, respectively. While we never ran systematic comparisons, our informal assessment was that these methods generally did not improve performance relative to our current approach and sometimes reduced it.}

\subsubsection{Comparing the Coding Schemes}

This section has discussed three information-theoretic approaches to multitask regression: RIC, Full MIC, and Partial MIC. In general, the negative log-likelihood portion of RIC may differ from that of the other two, because Full and Partial MIC may use a nondiagonal covariance estimate like $\widehat{\Sigma}_F$, while RIC, only operating on one response at a time, implicitly uses $\widehat{\Sigma}_D$. However, since we also use $\widehat{\Sigma}_D$ for Full and Partial MIC, the real differences come from the coding penalties.

These are compared in Table \ref{compare} for various values of $k$, the number of responses for which we add the current feature under consideration. Of course, Full MIC is only allowed to take $k = 0$ or $k = h$, so it actually has $h$ nonzero coefficients in all three rows of the table. However, if the extra $h - k$ coefficients correspond to non-predictive features, the extra reduction in residual-coding cost that Full MIC enjoys over the other methods is likely to be small.

The numbers beside the coding-cost formulas illustrate the case $m = 2{,}000$ features and $h = 20$ responses. As expected, each coding scheme is cheapest in the case for which it was designed; however, we note that the MIC methods are never excessively expensive, unlike RIC for $k = h$.

\begin{table*} [htbp]
\caption{Costs in bits for each of the three schemes to code the appearance of a feature in $k=1$, $k=\frac{h}{4}$, and $k=h$ response models. In general, we assume $m \gg h \gg 1$. Note that for $h \in \{5, \ldots, 1000\}$, $c_h \approx 1$. Examples of these values for $m=2{,}000$ and $h=20$ appear in brackets; the smallest of the costs appears in bold.}
\centering
\label{compare}
\begin{center}
\begin{tabular}{|c|cc|cc|cc|}
\toprule
$k$ & \multicolumn{2}{c|}{Partial MIC} & \multicolumn{2}{c|}{Full MIC} & \multicolumn{2}{c|}{RIC}\\
\midrule
1 & $\lg m + c_h + \lg h + 2$ & [18.4]  & $\lg m + 2 h$ & [51.0] & $\lg m+ 2$ & {\bf [13.0]}\\
$\frac{h}{4}$ & $\lg m + \lg^*\left (\frac{h}{4} \right) + c_h + \lg {h \choose h/4} + \frac{h}{2}$ & {\bf [39.8]} & $\lg m + 2 h$ & [51.0] & $\frac{h}{4} \lg m + \frac{h}{2}$ & [64.8]\\
$h$ & $\lg m + \lg^* h + c_h + 2 h$ & [59.7] & $\lg m + 2 h$ & {\bf [51.0]}  & $h \lg m + 2 h$ & [259.3]\\
\bottomrule
\end{tabular}
\end{center}
\end{table*}

\subsection{Implementation}\label{reg-implementation}

As mentioned in Section \ref{Penalized Regression}, searching over all possible combinations of zero and nonzero coefficients for the model that minimizes description length is computationally intractable. We therefore implement MIC using a forward stepwise search procedure, detailed in Algorithm \ref{StepwiseRegressionMIC}. Beginning with a null model containing only an intercept feature, we evaluate each feature for addition to this model and add the one that would most reduce description length, as computed using Algorithm \ref{DescrLength}. Updating our model, we re-evaluate each remaining feature and add to the model the next-best feature. We continue until we find no more features that would reduce description length if added to the model.

\begin{pseudocode}[shadowbox]{StepwiseRegressionMIC}{X, Y, \text{method}}\label{StepwiseRegressionMIC}
\text{ // ``method" is either ``partially dependent MIC" or ``fully dependent MIC."}\\
\text{Make sure that the matrix $X$ contains a column of 1's for the intercept. }\\
\text{Initialize the model $\bhat$ to have nonzero coefficients only for the intercept terms. That is, if $\bhat$}\\
\text{\ \ \ is an $m \times h$ matrix, exactly the first row has nonzero elements.}\\
\text{Compute $\widehat{\Sigma}$ based on the residuals $Y - X \bhat$ (whether as $\widehat{\Sigma}_F$, $\widehat{\Sigma}_D$, or something in between).}\\
\WHILE \TRUE\\
\BEGIN
\text{Find the feature $f$ that, if added, would most reduce description length.}\\
\text{ \ \ \ (If method is ``fully dependent MIC," then ``adding a feature" means making}\\
\text{ \ \ \ nonzero the coefficients for all of the responses with that feature. If method is ``partially}\\
\text{ \ \ \ dependent MIC," then it means making nonzero the optimal subset of responses,}\\
\text{ \ \ \ where the ``optimal subset" is approximated by a stepwise-style search}\\
\text{ \ \ \ through response subsets.)}\\
\text{Let $\bhat_f$ denote the model that would result if this new feature were added.}\\
\IF \CALL{DescrLength}{X, Y, \bhat_f, \widehat{\Sigma}, \text{method}} < \CALL{DescrLength}{X, Y, \bhat, \widehat{\Sigma}, \text{method}}
\THEN
\BEGIN
\bhat \GETS \bhat_f\\
\text{Update $\widehat{\Sigma}$.}\\
\END
\ELSE \BREAK
\END\\
\RETURN \bhat
\end{pseudocode}

\begin{pseudocode}[shadowbox]{DescrLength}{X, Y, \bhat, \widehat{\Sigma}, \text{method}}\label{DescrLength}
\text{Let $m$ be the number of features (columns of $X$) and $h$ the number of responses (columns of $Y$).}\\
d \GETS 0 \ \ \ \text{ // Initialize $d$, the description length.}\\
d \GETS d + \mathcal{D}(Y \giv \bhat) \ \ \ \text{ // Add the likelihood cost using \eqref{fullcoveq}. }\\
\FOR j \GETS 1 \TO m \ \ \ \ \text{ // Add the cost of feature $j$, if any. }
\DO
\BEGIN
k \GETS \text{number of nonzero responses for feature $j$}\\
\IF k > 0 \THEN
\BEGIN
d \GETS d + \lg m \ \  \text{ // Cost to specify which feature.}\\
d \GETS d + 2 k \ \ \ \ \text{ // Cost to specify nonzero coefficients.}\\
\IF \text{method == ``partially dependent MIC"} \THEN
d \GETS d + \lg^* k + c_h + \lg \binom{h}{k} \ \ \ \text{ // Specify which subset of responses.}
\END
\END\\
\RETURN d
\end{pseudocode}

The above procedure requires evaluating the quality of features $\mathcal{O}(m r)$ times, where $r$ is the number of features eventually added to the model. For Partial MIC, each time we evaluate a feature, we have to determine the best subset of responses that might be associated with that feature. As Algorithm \ref{StepwiseRegressionMIC} notes, this is also done using a stepwise-style search: Start with no responses, add the best response, then re-evaluate the remaining responses and add the next-best response, and so on.\footnote{A stepwise search that re-evaluates the quality of each response at each iteration is necessary because, if we take the covariance matrix $\widehat{\Sigma}$ to be nondiagonal, the values of the residuals for one response may affect the likelihood of residuals for other responses. If we take $\widehat{\Sigma}$ to be diagonal, as we end up doing in practice, then an $\mathcal{O}(h)$ search through the responses without re-evaluation would suffice. } Unlike an ordinary stepwise algorithm, however, we don't terminate the search if, at the current number of responses in the model, the description length fails to be the best seen yet. Because we're interested in borrowing strength across responses, we need to avoid overlooking cases where the correlation of a feature with any single response is insufficiently strong to warrant addition, yet the correlations with all of the responses are. Moreover, the $\lg \binom{h}{k}$ term in Partial MIC's coding cost does not increase monotonically with $k$, so even if adding the feature to an intermediate number of response models doesn't look promising, adding it to all of them might. Thus, we perform a full $\mathcal{O}(h^2)$ search in which we eventually add all the responses to the model. As a result, Partial MIC requires a total of $\mathcal{O}(m r h^2)$ evaluations of description length.\footnote{Full MIC, not requiring the search through subsets of responses, is only $\mathcal{O}(m r)$ in the number of evaluations of description length.}

However, a few optimizations can reduce the computational cost of Partial MIC in practice.
\begin{itemize}
\item We can quickly filter out most of the irrelevant features at each iteration by evaluating, for each feature, the decrease in negative log-likelihood that would result from simply adding it with all of its responses, without doing any subset search. Then we keep only the top $t$ features according to this criterion, on which we proceed to do the full $\mathcal{O}(h^2)$ search over subsets. We use $t = 75$, but we find that as long as $t$ is bigger than, say, 10 or 20, it makes essentially no impact to the quality of results. This reduces the number of model evaluations to $\mathcal{O}(mr + r t h^2)$.
\item We can often short-circuit the $\mathcal{O}(h^2)$ search over response subsets by noting that a model with more nonzero coefficients always has lower negative log-likelihood than one with fewer nonzero coefficients. This allows us to get a lower bound on the description length for the current feature for each number $k \in \set{1, \ldots, h}$ of nonzero responses that we might choose as
\sn{boundEqn}{
&(\text{model cost for other features already in model}) \\
 &+ (\text{negative log-likelihood of $Y$ if all $h$ responses for this feature were nonzero})\\
& +  (\text{the increase in model cost if $k$ of the responses were nonzero}).
}
We then need only check those values of $k$ for which \eqref{boundEqn} is smaller than the best description length for any candidate feature's best response subset seen so far. In practice, with $h=20$, we find that evaluating $k$ up to, say, 3 or 6 is usually enough; i.e., we typically only need to add $3$ to $6$ responses in a stepwise manner before stopping, with a cost of only $3h$ to $6h$.\footnote{If $\widehat{\Sigma}$ is diagonal and we don't need to re-evaluate residual likelihoods at each iteration, the cost is only $3$ to $6$ evaluations of description length.}
\end{itemize}

Although we did not attempt to do so, it may be possible to formulate MIC using a regularization path, or \emph{homotopy}, algorithm of the sort that has become popular for performing $\ell_1$ regularization without the need for cross-validation (e.g., \cite{friedman2008rpg}). If possible, this would be significantly faster than stepwise search.

\subsection{Experiments}\label{Experiments-sec2}

We evaluate MIC on several synthetic and real data sets in which multiple responses are associated with the same set of features. We focus on the parameter regimes for which MIC was designed: Namely, $m \gg n$, but with only a relatively small number of features expected to be predictive. We describe the details of each data set in its own subsection below.

For comparing MIC against existing multitask methods, we used the Matlab ``Transfer Learning Toolkit" \cite{TLToolkit}, which provides implementations of seven algorithms from the literature. Unfortunately, most of them did not apply to our data sets, since they often required \emph{meta-features} (features describing other features), or expected the features to be frequency counts, or were unsupervised learners. The two methods that did apply were AndoZhang and BBLasso, both described in Section \ref{Multitask Learning}.

The AndoZhang implementation was written by the TL Toolkit authors, Wei-Chun Kao and Alexander Rakhlin. It included several free parameters, including the regularization coefficients $\lambda_k$ for each task (all set to 1, as was the default), the optimization method (also set to the default), and $H$. \cite[pp. 1838-39]{ando2005flp} tried values of $H$ between 10 and 100, settling on 50 but finding that the exact value was generally not important. We performed informal cross-validation for values between 1 and 250 and found, perhaps surprisingly, that $H = 1$ consistently gave the best results. We used this value throughout the experiments below.

The BBLasso implementation was written by Guillaume Obozinski, based on a paper \cite{obozinski2006mtf} published earlier than the \cite{obozinski} cited above; however, the algorithm is essentially the same. For each parameter, we used the default package setting.

AndoZhang and BBLasso are both classification methods, so in order to compare against them, we had to turn MIC into a classification method as well. One way would have been to update \eqref{fullcoveq} to reflect a logistic model; however, the resulting computation of $\bhat$ would then have been nonlinear and much slower. Instead, we continue to apply \eqref{fullcoveq} and simply regress on responses that have the values 0 or 1. Once we've chosen which features will enter which models, we do a final round of logistic regression for each response separately with the chosen features to get a slightly better classifier.


\subsubsection{Synthetic Data}\label{Synthetic Data}

We created synthetic data according to three separate scenarios, which we'll call ``Partial," ``Full," and ``Independent." For each scenario, we generated a matrix of continuous responses as
\begin{equation}
\nonumber
Y_\text{sim} = X_\text{sim} {\bf \beta}_\text{sim} + \epsilon_\text{sim},
\end{equation}
with $m=2{,}000$ features, $h=20$ responses, and $n=100$ observations. Then, to produce binary responses, we set to 1 those response values that were greater than or equal to the average value for their columns and set to 0 the rest, yielding a roughly 50-50 split between 1's and 0's because of the normality of the data.
Each entry of $X_\text{sim}$ was i.i.d. $\mathcal{N}(0,1)$, each nonzero entry of $\beta_\text{sim}$ was i.i.d. $\mathcal{N}(0,1)$, and entry of $\epsilon_\text{sim}$ was i.i.d. $\mathcal{N}(0, 0.1)$, with no covariance among the $\epsilon_\text{sim}$ entries for different responses.\footnote{We mentioned in Section \ref{logLikeMultipleY} that MIC performed best when we used $\widehat{\Sigma}_D$, the diagonal covariance-matrix estimate. One might wonder whether this was due only to the fact that we created our synthetic test data without covariance among the responses. However, this was not the case. When we generated synthetic noise using a correlation matrix in which each off-diagonal entry was a specific nonzero value (in particular, we tried 0.4 and 0.8), the same trend appeared: MIC with $\widehat{\Sigma}_D$ had slightly lower test error.} Each response had $4$ beneficial features, i.e., each column of $\beta_\text{sim}$ had 4 nonzero entries.

\begin{figure}[h]
\begin{minipage}{0.3\linewidth}
\centering
\includegraphics[scale=.35]{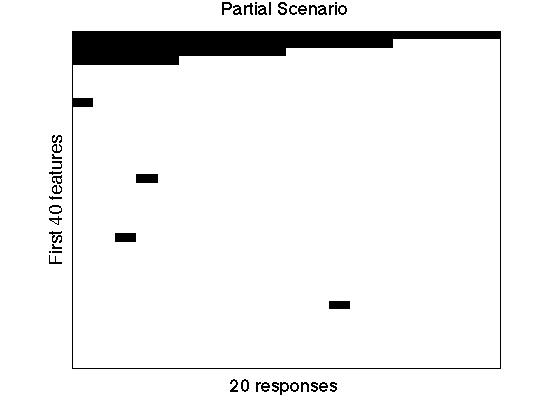}
\label{17JanSc1}
\end{minipage}
\hspace{0.015\linewidth} 
\begin{minipage}{0.3\linewidth}
\centering
\includegraphics[scale=.35]{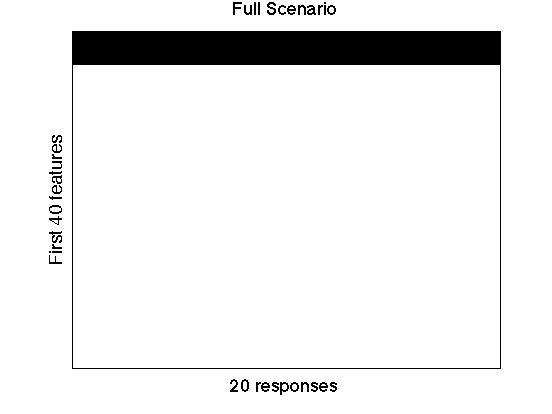}
\label{17JanSc2}
\end{minipage}
\hspace{0.015\linewidth} 
\begin{minipage}{0.3\linewidth}
\centering
\includegraphics[scale=.35]{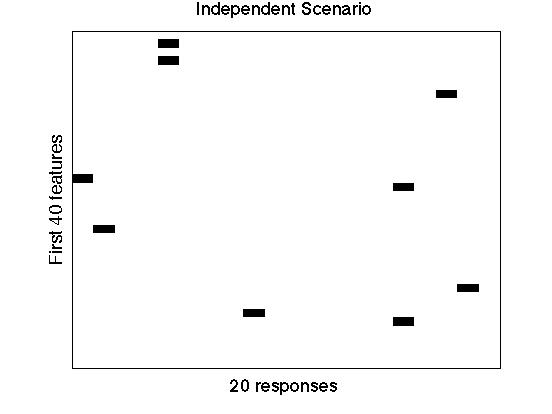}
\label{17JanSc3}
\end{minipage}
\caption{Examples of $\beta_\text{sim}$ for the three scenarios, with the nonzero coefficients in black. The figures show all 20 columns of $\beta_\text{sim}$ but only the first 40 rows. (For the Partial and Independent scenarios, the number of nonzero coefficients appearing in the first 40 rows is exaggerated above what would be expected by chance for illustration purposes.)}
\label{sampleBetas}
\end{figure}

The scenarios differed according to the distribution of the beneficial features in $\beta_\text{sim}$.
\begin{itemize}
\item In the Partial scenario, the first feature was shared across all 20 responses, the second was shared across the first 15 responses, the third across the first 10 responses, and the fourth across the first 5 responses. Because each response had four features, those responses ($6-20$) that didn't have all of the first four features had other features randomly distributed among the remaining features (5, 6, \ldots, 2000).
\item In the Full scenario, each response shared exactly features $1-4$, with none of features $5-2000$ being part of the model.
\item In the Independent scenario, each response had four random features among $\set{1, \ldots, 2000}$.
\end{itemize}
Figure \ref{sampleBetas} illustrates these feature distributions, showing the first 40 rows of random $\beta_\text{sim}$ matrices.

Table \ref{synthetic-sec2} shows the performance of each of the methods on five random instances of these data sets. Test-set errors for the ``True Model" are those obtained with logistic regression for each response separately using a model containing exactly the features of $\beta_\text{sim}$ that had nonzero coefficients. In addition to 
test-set error, we show precision and recall metrics. \emph{Coefficient-level} precision and recall refer to individual coefficients: For those coefficients of the data-generating $\beta_\text{sim}$ that were nonzero, did our final $\bhat$ show them as nonzero, and vice versa? \emph{Feature-level} precision and recall look at the same question for entire features: If a row of $\beta_\text{sim}$ had \textit{any} nonzero coefficients, did our corresponding row of $\bhat$ have any nonzero coefficients (not necessarily the same ones), and vice versa?

The best test-set error values are bolded (though not necessarily statistically significant). As we would expect, each of Partial MIC, Full MIC, and RIC is the winner in the synthetic-data regime for which it was designed, although Partial MIC competes with Full MIC even in the Full regime. Though the results are not shown, we observed informally that AndoZhang and BBLasso tended to have larger differences between training and testing error than MIC, implying that MIC may be less likely to overfit.  Resistance to overfitting is a general property of MDL approaches, because the requirement of completely coding $\bhat$ constrains model complexity. This point can also be seen from the high precision of the information-theoretic methods, which, surprisingly, show comparable recall to BBLasso. Full MIC behaves most like BBLasso, because both algorithms select entire rows of coefficients for addition to $\bhat$.

\begin{table*}[t]
\caption{Test-set accuracy, precision, and recall of MIC and other methods on 5 instances of the synthetic data sets generated as described in Section \ref{Synthetic Data}. Because synthetic data is cheap and our algorithms, once trained, are fast to evaluate, we used $10{,}000$ test data points for each data-set instance. Standard errors are reported over each task; that is, with 5 data sets and 20 tasks per data set, the standard errors represent the sample standard deviation of 100 values divided by $\sqrt{100}$ (except for feature-level results, which apply only to entire data sets and so are divided by $\sqrt{5}$). Baseline accuracy, corresponding to guessing the majority category, is roughly 0.5. AndoZhang's NA values are due to the fact that it does not explicitly select features.}
\centering
\begin{tabular}{ccccccc}
\toprule
Method & True Model & Partial MIC & Full MIC & RIC & AndoZhang & BBLasso\\ 
\midrule
\multicolumn{7}{c}{Partial Synthetic Data Set}\\
\midrule
Test error
 & $0.07 \pm 0.00$
&{\bf  0.10 $\pm$ 0.00}
 & $0.17 \pm 0.01$
  & $0.12 \pm 0.01$
 & $0.50 \pm 0.02$
 & $0.19 \pm 0.01$
\\ 
Coeff. precision 
& $1.00 \pm 0.00$
 & $0.84 \pm 0.02$
 & $0.26 \pm 0.01$
 & $0.84 \pm 0.02$
 & NA
 & $0.04 \pm 0.00$
\\ 
Coeff. recall
& $1.00 \pm 0.00$
 & $0.77 \pm 0.02$
 & $0.71 \pm 0.03$
 & $0.56 \pm 0.02$
 & NA
 & $0.81 \pm 0.02$
\\ 
Feature precision
& $1.00 \pm 0.00$
 & $0.99 \pm 0.01$
 & $0.97 \pm 0.02$
  & $0.72 \pm 0.05$
 & NA
 & $0.20 \pm 0.03$
\\ 
Feature recall
& $1.00 \pm 0.00$
 & $0.54 \pm 0.05$
 & $0.32 \pm 0.03$
  & $0.62 \pm 0.04$
 & NA
 & $0.54 \pm 0.01$
\\ 
\midrule
\multicolumn{7}{c}{Full Synthetic Data Set}\\
\midrule
Test error
& $0.07 \pm 0.00$
 &{\bf  0.08 $\pm$ 0.00}
 &{\bf 0.08 $\pm$ 0.00}
  & $0.11 \pm 0.01$
 & $0.45 \pm 0.02$
 & $0.09 \pm 0.00$
\\ 
Coeff. precision
& $1.00 \pm 0.00$
 & $0.98 \pm 0.01$
 & $0.80 \pm 0.00$
  & $0.86 \pm 0.02$
 & NA
 & $0.33 \pm 0.03$
\\ 
Coeff. recall
&  $1.00 \pm 0.00$
 & $1.00 \pm 0.00$
 & $1.00 \pm 0.00$
  & $0.63 \pm 0.02$
 & NA
 & $1.00 \pm 0.00$
\\ 
Feature precision
& $1.00 \pm 0.00$
 & $0.80 \pm 0.00$
 & $0.80 \pm 0.00$
  & $0.36 \pm 0.06$
 & NA
 & $0.33 \pm 0.17$
\\ 
Feature recall
& $1.00 \pm 0.00$
 & $1.00 \pm 0.00$
 & $1.00 \pm 0.00$
  & $1.00 \pm 0.00$
 & NA
 & $1.00 \pm 0.00$
\\ 
\midrule
\multicolumn{7}{c}{Independent Synthetic Data Set}\\
\midrule
Test error
& $0.07 \pm 0.00$
 & $0.17 \pm 0.01$
 & $0.36 \pm 0.01$
 & {\bf 0.13 $\pm$ 0.01}
 & $0.49 \pm 0.00$
 & $0.35 \pm 0.01$
\\ 
Coeff. precision
& $1.00 \pm 0.00$
 & $0.95 \pm 0.01$
 & $0.06 \pm 0.01$
 & $0.84 \pm 0.02$
 & NA
 & $0.02 \pm 0.00$
\\ 
Coeff. recall
& $1.00 \pm 0.00$
 & $0.44 \pm 0.02$
 & $0.15 \pm 0.02$
 & $0.58 \pm 0.02$
 & NA
 & $0.43 \pm 0.02$
\\ 
Feature precision
& $1.00 \pm 0.00$
 & $1.00 \pm 0.00$
 & $1.00 \pm 0.00$
 & $0.83 \pm 0.02$
 & NA
 & $0.30 \pm 0.05$
\\ 
Feature recall
& $1.00 \pm 0.00$
 & $0.44 \pm 0.02$
 & $0.14 \pm 0.02$
 & $0.58 \pm 0.03$
 & NA
 & $0.42 \pm 0.06$
\\ 
\bottomrule
\end{tabular}
\label{synthetic-sec2}
\end{table*}

\subsubsection{Yeast Growth}\label{Yeast Growth}

Our first real data set comes from \cite[pp. 5-6]{litvin2009mig}. It consists of real-valued growth measurements of 104 strains of yeast ($n=104$ observations) under 313 drug conditions. In order to make computations faster, we hierarchically clustered these 313 conditions into 20 groups using single-link clustering with correlation as the similarity measure. Taking the average of the values in each cluster produced $h=20$ real-valued responses, which we then binarized into two categories: values at least as big as the average for that response (set to 1) and values below the average (set to 0).\footnote{The split was not exactly 50-50, as the data were sometimes skewed, with the mean falling above or below the median. Table \ref{yeastertel} reflects this fact, showing that a classifier that simply guesses the majority label achieves an average error rate of 0.41.} The features consisted of 526 markers (binary values indicating major or minor allele) and 6,189 transcript levels in rich media for a total of $m = 6{,}715$ features. 

The ``Yeast Growth" section of Table \ref{yeastertel} shows 
test errors from 5-fold CV on this data set. Though the difference isn't statistically significant, Partial MIC appears to outperform BBLasso on test error. In any event, Partial MIC produces a much sparser model, as can be seen from the numbers of nonzero coefficients and features: Partial MIC includes nonzero coefficients in an average of 4 rows of $\bhat$, in contrast to 63 for BBLasso. Partial MIC also appears to outperform Full MIC and RIC, presumably because the assumptions of complete sharing or no sharing of features across responses rarely hold for real-world data. Like Partial MIC, AndoZhang did well on this data set; however, the algorithm scales poorly to large numbers of responses and so took 39 days to run.\footnote{Much of the reason for this is that AndoZhang, as currently implemented, only produces predictions for one of the responses after it trains on all of them. Thus, in order to predict all 20 responses, we had to run it separately 20 times. It is probably possible to train once and then predict all 20 responses simultaneously, but we wanted to avoid introducing errors by changing the code.}

\subsubsection{Yeast Markers}

The Yeast Growth data set described above includes as features both markers and transcripts for 104 yeast strains. We can consider these as variables for prediction in their own right, without reference to the growth responses at all. In fact, regression of transcripts against markers is commonly done; it's known as ``expression quantitative trait loci" (eQTL) mapping \cite{kendziorski2006rsm}. Since the marker variables are already binary (major or minor allele at the given location), we decided to flip the problem around from the usual eQTL setup: Using the $m=6{,}189$ transcripts as features, we predicted a subset of 20 of the 526 markers (numbered 25, 50, 75, \ldots, 500 for a good distribution along the genome).

The results are shown in the ``Yeast Markers" section of Table \ref{yeastertel}. Unlike the case of the Yeast Growth data, there is apparently less sharing of feature information across markers, as RIC appears to outperform Partial MIC on test error. We did not run AndoZhang on this data set.

\begin{table*}[t]
\caption{Accuracy and number of coefficients and features selected on five folds of CV for the Yeast Growth, Yeast Markers, and Breast Cancer data sets. Standard errors are over the five CV folds; i.e., they represent (sample standard deviation) / $\sqrt{5}$.  The ``Majority Label" column represents a classifier which guesses the more common of the 0 / 1 labels seen in the training set. AndoZhang's NA values are due to the fact that it does not explicitly select features.}
\centering
\begin{tabular}{ccccccc} 
\hline
Method & Partial MIC  & Full MIC & RIC & AndoZhang & BBLasso & Majority Label\\ 
\midrule
\multicolumn{7}{c}{Yeast Growth}\\
\midrule
Test error 
 &{\bf  0.38 $\pm$ 0.04}
 & $0.39 \pm 0.04$
  & $0.41 \pm 0.05$
 & $0.39 \pm 0.03$
 & $0.43 \pm 0.03$
  & $0.41 \pm 0.04$
\\
Num. coeff. sel. 
  & $22 \pm 4$
 & $64 \pm 4$
 & $9 \pm 1$
 & NA
 & $1268 \pm 279$
  & NA
 \\
 Num. feat. sel. 
  & $4 \pm 0$
 & $3 \pm 0$
 & $9 \pm 1$
 & NA
 & $63 \pm 14$
  & NA
 \\
 \midrule
\multicolumn{7}{c}{Yeast Markers}\\
\midrule
Test Error
 & $0.34 \pm 0.07$
  & $0.44 \pm 0.04$
 & {\bf 0.27 $\pm$ 0.06}
& -
 & $0.45 \pm 0.05$
 & $0.53 \pm 0.03$
\\
Num. coeff. sel. 
 & $20 \pm 1$
 & $68 \pm 26$
 & $37 \pm 2$
& NA
 & $1044 \pm 355$
& NA 
\\
Num. feat. sel. 
 & $16 \pm 1$
 & $3 \pm 1$
 & $37 \pm 2$
& NA
 & $52 \pm 18$
 & NA
\\
\midrule
\multicolumn{7}{c}{Breast Cancer}\\
\midrule
Test error 
& {\bf  0.33 $\pm$ 0.08}
 & $0.37 \pm 0.08$
  & $0.36 \pm 0.08$
 & $0.44 \pm 0.03$
 &{\bf 0.33 $\pm$ 0.08}
  & $0.39 \pm 0.04$
\\
Num. coeff. sel. 
  & $3 \pm 0$
 & $11 \pm 1$
  & $2 \pm 0$
 & NA
 & $61 \pm 19$
  & NA
 \\
 Num. feat. sel. 
  & $2 \pm 0$
 & $2 \pm 0$
  & $2 \pm 0$
 & NA
 & $12 \pm 4$
  & NA
 \\
\bottomrule
\end{tabular}
\label{yeastertel}
\end{table*}

\subsubsection{Breast Cancer}

Our Breast Cancer data set represents a combination of five of the seven data sets used in \cite{vantveer}. It contains  $1{,}171$ observations for $22{,}268$ RMA-normalized gene-expression values. We considered five associated responses; two were binary---prognosis (``good" or``poor") and ER status (``positive" or ``negative")---and three were not---age (in years), tumor size (in mm), and grade (1, 2, or 3). We binarized the three non-binary responses into two categories: Response values at least as high as the average, and values below the average. Some of the responses were unavailable for some observations, so we eliminated those observations, leaving 882. Of those, we kept only the first $n=100$, both to save computational resources and to make the problem ``harder." To reduce the features to a manageable number, we took the $m=5{,}000$ that had highest variance.

Table \ref{yeastertel} shows the results. In this case, BBLasso did as well as Partial MIC on test error; however, as usual, Partial MIC produced a much sparser model. Sparsity is important for biologists who want to interpret the selected features.

\section{Hypothesis Testing with a Single Response}\label{hypothesis}

As the experimental results in Section \ref{Experiments-sec2} demonstrate, regression can be an effective tool for predicting one or more responses from features, and in cases where the number of features is large, selection can improve generalization performance. However, as we noted with the Yeast and Breast Cancer data sets, accuracy is not always the only goal; sometimes model interpretability is important. Indeed, in science, the entire goal of regression is often not prediction but rather discovering a ``true model," by testing hypotheses about whether pairs of variables really have a linear relationship. For instance, an econometrician regressing housing prices on air-pollution levels and resident income \cite{harrison1978hhp} is not trying to predict price levels but to study consumer demand behavior.

Hypothesis testing with a small number of features is straightforward: Given a regression coefficient, we can compute a $t$ statistic, and if it exceeds some threshold, we reject the null hypothesis that the coefficient is 0. The situation becomes slightly more complicated when we have many features whose coefficients we want to examine. Section \ref{previous-approaches} reviews some standard approaches to this multiple-testing problem, and Section \ref{MDL and Likelihood-Ratio Tests} recasts them in the light of MDL. This sets the stage for an MIC approach to hypothesis testing in Section \ref{hypothesis-mic}.

\subsection{Multiple-Testing Procedures}\label{previous-approaches}

Consider the case of a single response variable $y$ (say, a gene-transcript level), and suppose we're trying to determine which of $m$ features (say, genetic markers) are linearly correlated with the response. If we test each feature at a significance level $\al$, the overall probability that we'll falsely reject a true null hypothesis will be much greater than $\al$---this is the problem of \emph{multiple hypothesis testing}.

\subsubsection{Bonferroni Correction}

A standard way to control against so-called \emph{alpha inflation} is called the \emph{Bonferroni correction}: Letting $H_1, \ldots, H_m$ denote the null hypotheses and $p_1, \ldots, p_m$ the associated p-values, we reject $H_j$ when $p_j \leq \frac{\alpha}{m}$. By Boole's inequality, this controls what is known as the \emph{family-wise error rate} (FWER), the probability of making any false rejection, at level $\alpha$ under the \emph{complete null hypothesis} that all of $H_1, \ldots, H_m$ are true.\footnote{When the test statistics are independent or \emph{positive orthant dependent}, one can replace $\frac{\alpha}{m}$ by $1 - (1-\alpha)^\frac{1}{m}$, but the resulting improvement in power is small for small $\alpha$ \cite[p. 570]{shaffer1995mht}.} In fact, Bonferroni controls FWER in a \emph{strong sense} as well: For any subset of null hypotheses, the probability of falsely rejecting any member of that subset when all members are true is also bounded by $\alpha$ \cite[p. 800]{hochberg1988sbp}.

The Bonferroni correction is a \emph{single-stage testing procedure} in which, unfortunately, testing a larger number of hypotheses reduces the power for rejecting any one of them \cite[p. 569]{shaffer1995mht}. This is especially problematic when the number of hypotheses tested is in the thousands or tens of thousands, as is common with, e.g., microarray or fMRI data. However, statisticians have developed several \emph{multistage testing procedures} that help to overcome this limitation by conditioning rejection thresholds for some test statistics on the results of rejections by others.

\subsubsection{Improvements on Bonferroni}

One of the first of these was the \emph{Holm step-down procedure} \cite{holm1979ssr}. Here, we let $p_{(1)}, \ldots, p_{(m)}$ denote the p-values sorted in increasing order and $H_{(1)}, \ldots, H_{(m)}$ the corresponding null hypotheses. We begin by looking at $p_{(1)}$: If it fails to be $\leq \frac{\alpha}{m}$, we stop without rejecting anything. (This is what Bonferroni would do as well, since no p-value is $\leq \frac{\alpha}{m}$.) However, if we do reject $H_{(1)}$, we move on to $H_{(2)}$ and reject if and only if $p_{(2)} \leq \frac{\alpha}{m-1}$. We continue in this manner, rejecting $H_{(j)}$ when $p_{(j)} \leq \frac{\alpha}{m - j + 1}$, until we fail to reject a hypothesis. Like Bonferroni, the Holm method controls FWER in the strong sense for independent or dependent test statistics \cite[p. 800]{hochberg1988sbp}.

Simes \cite[p. 751]{simes1986ibp} proposed a more lenient approach: Reject the complete null hypothesis if \textit{any} of the following inequalities holds, $j = 1, 2, \ldots, m$:
\s{
p_{(j)} < \f{j \al}{m}.
}
This controls FWER  for independent test statistics at level $\alpha$. Moreover, simulation studies suggested that it remained true for various multivariate-normal and multivariate-gamma test statistics \cite[p. 752]{simes1986ibp}. Unfortunately, unlike the Bonferroni and Holm procedures, the Simes approach says nothing about rejecting individual hypotheses once the complete null is rejected \cite[p. 754]{simes1986ibp}, although \cite{hochberg1988sbp} and \cite{hommel1988srm} subsequently proposed limited procedures for doing so.

\subsubsection{FDR and the BH Procedure}

In his closing discussion, Simes \cite[p. 754]{simes1986ibp} proposed that if we wanted to reject individual null hypotheses, we might reject the ordered hypotheses $H_{(1)}, \ldots, H_{(q)}$ such that
\sn{BH-procedure}{
q = \max \set{j : p_{(j)} \leq \f{j \al}{m} }.
}
However, he cautioned that there was ``no formal basis" for this. Indeed, \cite[p. 384]{hommel1988srm} showed that in certain cases, the FWER of this procedure could be made arbitrarily close to 1 for large $m$.

However, Benjamini and Hochberg \cite{benjamini_controlling_1995} pointed out that the Simes procedure could be said to control a different measure, which they called the \emph{false-discovery rate} (FDR). Letting $V$ denote the (unobservable) random variable for the number of true null hypotheses rejected and $S$ the (unobservable) random variable for the number of correctly rejected null hypotheses,
\s{
\text{FDR} := P(V+S > 0) \ E\brac{\f{V}{V+S} \Giv V + S > 0}
}
(where $V+S$, the number of null  hypotheses rejected in total, is observable). Thus, FDR is the expected proportion of falsely rejected null hypotheses. This statistic is more flexible than FWER because it accounts for the total number of hypotheses considered; if, for instance, we had $m=1{,}000{,}000$ hypotheses, the FWER would be very high, but the \emph{proportion} of false rejections could still be low. Because of this flexibility for large $m$, FDR has become standard in bioinformatics, neuroscience, and many other fields.

\cite[p. 293]{benjamini_controlling_1995} showed that the \emph{Benjamini-Hochberg (BH) step-up procedure} \eqref{BH-procedure} controls FDR at $\al$ for independent test statistics and any configuration of false null hypotheses. \cite{benjamini2001cfd} extended this result to show that the procedure also controlled FDR for certain types of positive dependency. This included positively correlated and normally distributed one-sided test statistics, which occur often in, for instance, gene-expression measurements \cite[p. 370]{reiner2003ide}.

Various extensions to the BH procedure have been proposed. For instance, \cite{benjamini1999sdm} suggested a step-down approach for independent test statistics that, while not dominating the BH step-up procedure, tended experimentally to yield higher power.  \cite[p. 1169]{benjamini2001cfd} showed that replacing $\alpha$ by $\alpha / \sum_{j=1}^m \frac{1}{j}$  would control FDR at $\alpha$ for any test-statistic correlation structure, though this is often more conservative than necessary (p. 1183). \cite{yekutieli1999rbf} proposed a resampling approach for estimating FDR distributions based the data, in order to gain increased power when the test statistics were highly correlated. Many further modifications to the BH procedure have been put forward since. However, in this thesis, we stick with the original version given by \eqref{BH-procedure}.

\subsection{Hypothesis Testing by MDL}\label{MDL and Likelihood-Ratio Tests}

MDL compares log-likelihoods of data under various models, choosing a more complicated model if and only if it has sufficiently lower negative log-likelihood. Appendix A shows that, in fact, this process is equivalent to a standard statistical procedure known as a generalized likelihood-ratio test, at some implied significance level $\al$.

Hypothesis testing differs slightly from regression, however. The goal with regression is to minimize test-set error by choosing a few highly informative features. Because MIC penalizes for each feature, if we find several, very similar features that are all correlated with the response, stepwise regression by MIC will likely include only one of them. For instance, suppose the true model is $y = f_1 + 2 f_2 + \eps$, where $f_1$ and $f_2$ are nearly identical features. Not wanting to waste model-coding bits, regression MIC would probably give the model $y = 3 f_1 + \eps$ or $y = 3 f_2 + \eps$.

With hypothesis testing, we aim to find all of the features that are significantly correlated with the response. In the above example, we would hope to include both $f_1$ and $f_2$ in our set of relevant features. To do this, we regress $Y$ on each feature in isolation; we keep those features that, according to MDL, deserve nonzero regression coefficients. Conceptually, we can think of this as a for-loop over features, in a given iteration of which we call an MIC-stepwise-regression function using the regular $Y$ matrix but with an $X$ matrix that contains only the current feature (and an intercept). There's one catch, though: Since our for-loop searches through $m$ potential features to add, we're effectively doing $m$ hypothesis tests, and we need to incorporate some sort of multiple-testing penalty. Below we describe a way to motivate such a penalty from an MDL perspective.

Section \ref{singleResponse} described a scenario to motivate ordinary MDL regression: Sender wanted to transmit data $Y$ to Receiver when both Sender and Receiver knew the value of an associated matrix $X$ of features. Now suppose instead that there are $m$ different Receivers, where Receiver $i$ knows the values only of the $i^\text{th}$ feature (i.e., the $i^\text{th}$ column of $X$). All the Receivers want to reconstruct $Y$, so Sender has to transmit messages to each of them telling them how to do it, possibly using the feature that they individually know. The messenger Hermes visits Sender and tells her some ground rules: Sender must transmit directly to each Receiver only information about how to reconstruct $Y$ given a model. All model information itself (i.e., all the $\bhat$ coefficients) Sender has to transmit to Hermes, who will then visit each Receiver and tell him his appropriate coefficient (possibly 0 if Sender hasn't bothered to encode a model for that feature). As a bonus, Sender is allowed to include in each model the intercept coefficient of $Y$ for free, meaning that if Sender doesn't specify a feature regression coefficient for Receiver $i$, that Receiver at least gets the average $\overline{Y}$ to use in reconstructing $Y$. Letting $\bhat_q$ denote the model Sender tells Hermes, Sender's total description length is
\s{
\mathcal{D}(\bhat_q) + \mathcal{D}^m(Y \giv \bhat_q),
}
where
\sn{dm}{
\mathcal{D}^m(Y \giv \bhat_q) := \sum_{i = 1}^m \mathcal{D}(Y \giv \model_i).
}
Here, we define $\model_i$ to be a null model, containing only the free intercept term, if Sender doesn't specify a nonzero regression coefficient for Receiver $i$ in $\bhat_q$. Otherwise, $\model_i$ contains both a costless intercept term and the regression coefficient for feature $i$.

The result should be something like a series of hypothesis tests, in which the features in $\bhat_q$ correspond to the rejected null hypotheses. The idea is that, because Sender has to tell Hermes which coefficients go to which Receiver, the coding of $\bhat_q$ will involve a penalty that grows with the number of possible features, which is what we need to protect against alpha inflation from multiple tests.

\subsection{MDL Multiple-Testing Corrections}\label{mdl-corrections}

We consider two coding schemes that Sender might use to tell Hermes which features are contained in her model $\bhat_q$.

\subsubsection{Bonferroni-Style Penalty}

Suppose $m=5{,}000$ and Sender wants to use nonzero coefficients for features 226, 1,117 and 3,486. One way to tell Hermes this is to transmit the binary representation of each of these numbers, using an idealized $\lg m$ bits for each one. Then, as in Section \ref{codeModel}, Sender spends 2 bits to specify each coefficient. The resulting message to Hermes costs $3 \lg m + 3 \cdot 2$ bits. In general, Sender's description length will be
\sn{minimize-bonf}{
q \lg m + 2 q + \mathcal{D}^m(Y \giv \bhat_q).
}

For each feature $j$, let $- \lg \Lambda_j$ denote the decrease in residual coding cost that would result from using a nonzero coefficient for feature $j$. (See the Appendix for the motivation behind this notation.) Then minimizing \eqref{minimize-bonf} is equivalent to the following decision rule: Looking at the features in decreasing order of their $- \lg \Lambda_j$ values, add feature $j$ to the model as long as
\sn{bonf-decision}{
- \lg \Lambda_j \geq \lg m + 2.
}
Interestingly, \eqref{bonferroni-lgm} in Appendix \ref{bonferroni-appendix} shows that this is equivalent to a Bonferroni correction at some implied $\alpha$. In fact, according to Appendix \ref{closeness-approx}, taking a cost per coefficient around 2,\footnote{2.77 to be exact, for $m=1$ and one degree of freedom, though this changes somewhat with $m$ and the number of degrees of freedom.} as we did for stepwise regression, corresponds to an $\alpha$ around 0.05.

\subsubsection{BH-Style Penalty}\label{BH-Style Penalty}

If Sender expects to code more than one coefficient, rather than coding an entire index to specify each feature, she may find it advantageous to use a scheme similar to \eqref{coding-k}. There, the scheme was used to convey a subset of nonzero response coefficients in Partial MIC; here it would describe a subset of features from $\set{1, \ldots, m}$. Using this code, Sender's description length using $\bhat_q$ would be
\sn{BH-mdl-eqn}{
\lg^* q + c_m + \lg \binom{m}{q} + 2 q + \mathcal{D}^m(Y \giv \bhat_q).
}
Apart from the two (small) terms at the beginning, this can be made to correspond at some implied $\alpha$\footnote{In fact, the same $\alpha$ as for the Bonferroni case above.} with \eqref{BHwithlambda4} and thus approximately with the ordinary BH procedure.

\section{MIC for Hypothesis Testing}\label{hypothesis-mic}

As was the case with regression, real-world hypothesis-testing problems often involve multiple responses.  \cite[sec. 1]{turlach2005svs} discuss the task of identifying a subset of 770 wavelength features that best predict 14 chemical quantities (e.g., pH, organic carbon, and total cations). \emph{Expression quantitative trait loci} (eQTL) mapping is the process of looking for correlations between potentially thousands of gene-expression transcripts and genetic markers (e.g., \cite{schadt_genetics_2003}). Section \ref{Existing Approaches} reviews some existing ways to deal with this problem, and then Section \ref{The MIC Approach} describes the MIC approach. Experimental comparisons of the various methods are described in Section \ref{Experiments-hyp}.

\subsection{Existing Approaches}\label{Existing Approaches}

We describe a few examples of ways to perform hypothesis testing with multiple responses.

\subsubsection{Bonferroni and BH}

Probably the most straightforward approach is to treat each response separately and apply, say, the Bonferroni or BH procedures of Section \ref{previous-approaches} one response at a time. We'll refer to these approaches as simply ``Bonferroni" and ``BH," respectively.

According to \cite[p. 20]{kendziorski_statistical_2006}, a number of early eQTL studies took this approach, applying single-transcript mapping methods to each transcript separately. However, because such methods are designed to control false positives when analyzing a single transcript, multiple tests across transcripts may result in inflated FDR \cite[p. 23]{kendziorski_statistical_2006}.

\subsubsection{BonferroniMatrix and BHMatrix}

To account for the greater number of hypothesis tests being performed with $h > 1$ responses, we might penalize not just by $\frac{\alpha}{m}$ but $\frac{\alpha}{mh}$. The procedure that consists in applying the latter Bonferroni threshold to each p-value in our $m \times h$ matrix of p-values is what we'll call ``BonferroniMatrix." Effectively, we're imagining our matrix of p-values as one long vector and applying the standard Bonferroni correction to that.

Similarly, we can imagine turning our matrix of p-values into a single vector and applying the BH method to it: We'll call this approach ``BHMatrix." In contrast to BH, it has a harsher starting penalty ($\frac{\alpha}{mh}$ rather than $\frac{\alpha}{m}$), but it also allows us to pick out the best p-values from any location in the entire matrix, not just p-values in the particular column (response) that we're currently examining. \cite[p. 21]{kendziorski_statistical_2006} describes a conceptually similar approach, called ``Q-ALL," that used so-called q-values \cite{storey2003pfd} to identify significant marker-transcript correlations.

The result of BonferroniMatrix at level $\alpha$ is the same as that of regular Bonferroni at level $\frac{\alpha}{h}$, so if we're looking at a range of significance levels, it suffices to examine just Bonferroni or BonferroniMatrix; we'll show results for Bonferroni only in what follows. The same is not true for BH vs. BHMatrix, so we report on both methods.

\subsubsection{Empirical Bayes}\label{Empirical Bayes}

The BHMatrix approach does not treat each response separately, since, for example, if one response has lots of low p-values that pass the harshest thresholds of the step-up procedure, it can leave easier thresholds for the other responses. But this doesn't really ``share information" across responses in a meaningful way.

One approach that does is called \emph{EBarrays} \cite{kendziorski2003peb}, which uses an empirical Bayes hierarchical model to estimate a prior for the underlying means of each transcript (response), allowing for stable inference across transcripts despite the small sample size of each individually. The goal is to determine, at each marker (feature), which transcripts show significant association. Originally designed for sharing transcript information one marker at a time, the method has been extended to a \emph{mixture over markers} (MOM) model that uses the EM algorithm to assign probabilities that a transcript correlates to each marker \cite{kendziorski_statistical_2006}. As the names suggest, these methods apply specifically to eQTL problems, though of course, the empirical-Bayes shrinkage framework applies more generally. We give this as just an example of previous approaches to sharing strength across responses that have been developed for hypothesis testing. 

\subsection{The MIC Approach}\label{The MIC Approach}

While the EBarrays approach shares information across responses, it fails to take advantage of one potentially important source of shared strength: Namely, that features strongly associated with one response are perhaps more likely to be associated with other responses. In the case of eQTL, for example, we might suspect this type of sharing across responses by marker features corresponding to \textit{trans}-regulatory elements, which affect expression of many different genes.

On the other hand, the MIC approach of Section \ref{regression-section} does pick up on sharing of features across responses, so we propose a method for hypothesis testing by MIC. Just as the single-response version of hypothesis testing by MDL in Section \ref{MDL and Likelihood-Ratio Tests} applied single-response regression to each feature individually, so we can also imagine applying the multiple-response MIC regression approach of Section \ref{Information-Theoretic Regression---Many Responses} to each feature individually in order to do multiple-response hypothesis testing.

This amounts basically to a for-loop of the MIC Stepwise Regression Algorithm \ref{StepwiseRegressionMIC} over each feature, except that the costs of coding the features are different from usual because of the setup that Hermes imposed (see Section \ref{MDL and Likelihood-Ratio Tests}). Just as in that section, we can consider a Bonferroni-style (``Bonferroni-MIC") and a BH-style (``BH-MIC") coding scheme, using feature penalties of $q \lg m$ and $\lg^* q + c_m + \lg \binom{m}{q}$, respectively. Below we make each of these approaches more explicit.

\subsubsection{Implementation}\label{mic-hyp-implementation}

With Bonferroni-MIC, we can evaluate each feature one at a time for inclusion based on whether \textsc{StepwiseRegressionMIC} (Algorithm \ref{StepwiseRegressionMIC}) would have included nonzero coefficients for that feature, with the following exception. That algorithm includes a feature penalty of $\lg$ of the number of features (columns) in the given $X$ matrix. Here, when we give the algorithm only one column of $X$ at a time, this term would normally be just $\lg 1$. However, the coding scheme that Hermes imposed requires us instead to use $\lg m$ bits to specify a feature, where $m$ is the total number of features in $X$. Pseudocode appears in Algorithm \ref{BonferroniMICalg}.

\begin{pseudocode}[shadowbox]{Bonferroni-MIC}{X, Y, \text{method}}
\label{BonferroniMICalg}
\bhat \GETS 0\\
\FOR i = 1 \TO m \text{ \ \ \ // Loop over the features.}\\
\BEGIN
\text{\ Essentially, we call }\\
\ \ \ \ \bhat_i \GETS \CALL{StepwiseRegressionMIC}{X_i, Y, \text{method}},\\
\text{ \ except that the usual penalty term corresponding to $\lg$ of the number of columns of the }\\
\text{ \ input $X_i$ matrix (just 1 here) is replaced by $\lg m$, where $m$ is the number of columns of $X$.}\\
\END\\
\RETURN \bhat
\end{pseudocode}

BH-MIC is slightly trickier, because the cost of adding a new feature depends on how many features are already in the model. Our approach, outlined in Algorithm \ref{BH-MIC}, is to loop through features and evaluate whether they would include nonzero coefficients when no feature-coding cost is imposed (i.e., no $(\lg m)$-like term is charged). This fills $\bhat$ with a possibly inflated number of nonzero coefficients that we subsequently trim down if necessary. To do this, we evaluate the cost of keeping the best $q$ of the currently nonzero features for each $q$ from 0 to the current number of nonzero features $q_\text{orig}$, choose the optimal value $q^*$, and zero out the rest. Note that zeroing out those rows of $\bhat$ is sufficient; we have no need to go back and recompute the optimal subset of responses for the remaining features, because the cost to specify the included features is the same regardless of the response subset.

\begin{pseudocode}[shadowbox]{BH-MIC}{X, Y, \text{method}}\label{BH-MIC}
\bhat \GETS 0\\
\FOR i = 1 \TO m \text{ \ \ \ // Loop over the features.}\\
\BEGIN
\text{\ Essentially, we call }\\
\ \ \ \ \bhat_i \GETS \CALL{StepwiseRegressionMIC}{X_i, Y, \text{method}},\\
\text{ \ but without any cost to code features (the usual $\lg m$ term).  We also record $s_i$,}\\
\text{ \ the (positive) number of bits saved by using the current subset of nonzero coefficients $\bhat_i$ }\\
\text{ \ instead of having all the coefficients zero. If $\bhat_i == 0_{1 \times h}$, i.e., all the coefficients}\\
\text{ \ are already all zeros, $s_i \GETS 0$.}\\
\END\\
\text{Sort the $s_i$ in decreasing order and call them $s_{(i)}$. $s_{(1)}$ corresponds to saving the most bits, etc.}\\
q_\text{orig} \GETS \text{number of nonzero features currently in $\bhat$ (possibly too many)}\\
q^* \GETS \displaystyle \max_{q \in \set{1, \ldots q_\text{orig}}} \set{ \sum_{i=1}^q s_{(i)} - \paren{ \lg^* q + c_m + \lg \binom{m}{q} } } \text{ (or } q^* \GETS 0 \text{ if none of these is positive)} \\
\text{Zero out the rows of $\bhat$ corresponding to features whose inclusion saves fewer than $s_{(q^*)}$ bits.}\\
\RETURN \bhat
\end{pseudocode}

In both Bonferroni-MIC and BH-MIC, the algorithmic complexity is dominated by the for-loop. In Section \ref{reg-implementation}, we saw that a call to \textsc{StepwiseRegressionMIC} required $\mathcal{O}(m_i r h^2)$ evaluations of description length, where we've used $m_i$ to denote the number of columns of $X_i$. In particular, $m_i = 1$ for each $i$ and $r$ is at most 1, so the complexity within a given for-loop iteration is only $\mathcal{O}(h^2)$. Overall, then, these algorithms evaluate description length $\mathcal{O}(m h^2)$ times.

\subsubsection{Hypothesis-Testing Interpretation}\label{mic-hyp-interpretation}

Consider a single feature. In Section \ref{MDL and Likelihood-Ratio Tests}, we noted the correspondence between MDL regression on that feature and a statistical likelihood-ratio test of whether its regression coefficient was nonzero. With $h > 1$ responses, MIC hypothesis testing searches over subsets of responses to have nonzero coefficients with the feature. Each evaluation of description length during this process can be interpreted as a hypothesis test, in the way explained in the Appendix. In particular, when we evaluate the description length for some subset $S \subset \set{1, 2, \ldots, h}$ of the responses being nonzero, we're implicitly doing a likelihood-ratio test with these null and alternative hypotheses:
\sn{mult-h0-h1}{
&H_0: \beta = 0_{1 \times  h} \ \ \text{ vs. }\\
&H_1: \beta_i \neq 0 \text{ for $i \in S$,}
}
where $\beta$ is the $1 \times h$ matrix of true regression coefficients of the feature with the $h$ responses. In theory, MDL performs exponentially many such tests, one for each subset $S$,\footnote{Of course, in practice we do only $\mathcal{O}(h^2)$ of them during our stepwise-style search through responses to include.} though they are obviously highly correlated. We can interpret the $\lg^* k + c_h + \lg \binom{h}{k}$ penalty to specify the subset of $k$ of the responses (as described in Section \ref{multipleYModel}) as something of a multiple-testing correction for doing all of these implicit hypothesis tests.

Section \ref{mdl-corrections} and the corresponding portions of the Appendix pointed out the approximate correspondence between particular penalties and particular corrections: For instance, that $\lg m$ in log-likelihood space is essentially a Bonferroni correction in p-value space, or that a cost per coefficient of 2.77 approximately corresponds to $\alpha = 0.05$. With $h > 1$ responses, these approximations are somewhat rougher. The reason is that in \eqref{mult-h0-h1}, $k = \abs{S}$, the difference in dimension of the parameter spaces between $H_0$ and $H_1$, is often greater than 1. In this case, $- 2 \ln \Lambda \sim \chi^2_{(k)}$, and for $k > 1$, the standard-normal approximation using $\Phi$ in Appendix \ref{Example: Single Regression Coefficient} no longer goes through. Nevertheless, we can expect the Bonferroni- and BH-style penalties from Section \ref{mdl-corrections} to do roughly the right things for multiple responses; we needn't make them match the Bonferroni and BH procedures exactly. (If we did, what would be the point of using information theory?)

\subsection{Experiments}\label{Experiments-hyp}

When we measure the performance of our algorithms in terms of  test-set error, we can use cross-validation to run experiments on real data sets, as we did in Section \ref{Experiments-sec2}. When, instead, we need to judge how well our algorithm identifies truly nonzero coefficients, we have to fall back to synthetic data. That's because the real world doesn't tell us the true $\beta$ matrix\footnote{Indeed, there may not be any true $\beta$ matrix, because the data are not \emph{really} generated according to our model \eqref{modelsenderreceiver}, though the approximation works well enough.}---we only know it if we make it ourselves. Thus, in this section, we rely entirely on synthetic data. However, we can sometimes base our synthetic data on a real-world data set, and we do this with the Yeast Growth data in Section \ref{Simulated Yeast Data}.

For each of the scenarios below, we generated 25 random instances of the data sets, corresponding to 25 random $\beta_\text{sim}$ matrices, taking $n=100$ training data points and $h=20$ responses. Because synthetic data is cheap, we evaluate test-set error on $10{,}000$ test data points. We calculate error for each response as $\text{SSE} / \sqrt{n}$, where $\text{SSE}$ is the sum of squares error from the ordinary-least-squares regression of the given response on exactly the features selected by the algorithm (plus an intercept). We report precision and recall at the coefficient level (whether each particular nonzero coefficient was selected). We calculate these results separately for each response in each data set, so that standard errors represent standard deviations divided by $\sqrt{20 \cdot 25}$. 

We compare the following feature-selection approaches:
\begin{itemize}
\item ``Truth": Use an oracle to select exactly the true features.
\item ``Bonf,$\al$=$\al_0$": For each response separately, select those features whose p-values with the response are $\leq \al_0 / m$. We calculate the p-value for a given feature and response by regressing the response against only that feature and an intercept, and evaluating the p-value of the slope regression coefficient.
\item ``Indep,cpc=$c_0$": Apply the RIC algorithm described in Section \ref{Information-Theoretic Regression---Many Responses} to each response separately, using $c_0$ as the cost to code a coefficient value. ($c_0$ does not include the cost to specify which features enter the model, which is always $\lg m$. It shouldn't fall below 0, and the lowest value we tried was 0.1.)
\item ``Bonf-MIC": Bonferroni-style MIC, as described in Algorithm \ref{BonferroniMICalg}. The cost to code a coefficient is 2 bits.
\item ``BH,$\al$=$\al_0$": For each response separately, calculate p-values the same way as with ``Bonf,$\al$=$\al_0$," but apply the BH procedure \eqref{BH-procedure} at level $\al_0$ instead of the Bonferroni correction.
\item ``BHMat,$\al$=$\al_0$": Apply the BH procedure at level $\al_0$ to the entire $m \times h$ matrix of p-values all at once.
\item ``BH-MIC": BH-style MIC, as described in Algorithm \ref{BH-MIC}. The cost to code a coefficient is 2 bits.
\end{itemize}
The MIC methods have no free parameters (if we fix the cost per coefficient at 2 bits), so they give point values for precision and recall. To allow for comparison, then, we ran the other methods at a variety of $\al_0$ or $c_0$ levels, presenting results for $\al_0$ or $c_0$ levels at which precision approximately matched that of MIC; relative performance can then be assessed based on recall. Because we tried only a discrete grid of $\al_0$ or $c_0$ values, precisions did not always match exactly, so we took the highest precision not exceeding the MIC value. This puts MIC at a slight disadvantage in comparing recall, because its precision is often slightly higher than that of the other methods.

In our tables below, we make bold those results that represent the best performance among the three Bonferroni-style methods and also among the three BH-style methods. Here, ``best performance" is based on the observed average value, but the differences are often not statistically significant.

\subsubsection{Simulated Synthetic Data}\label{Simulated Synthetic Data}

We created three data sets in the same manner as described in Section \ref{Synthetic Data}, the only difference being that we took $m=1{,}000$ features instead of $2{,}000$ (for no particular reason). Table \ref{synthetic} shows the results for the Partial, Full, and Independent scenarios.

MIC appears to have had worse test-error than the other methods except in the Full scenario, where it performed close to the Truth. As the especially poor Independent performance suggests, this is probably due to the fact that MIC has a harder time picking up on single nonzero coefficients in a given row of $\beta_\text{sim}$, due to its ``higher overhead" cost to put them in (see Table \ref{compare} for $k=1$).

Of course, for hypothesis-testing algorithms, test error is not the most important metric. On recall, MIC does outperform its competitors in the Partial and especially Full scenarios. And as we would expect, it performs worse in the Independent case, again due to its high overhead for individual coefficients on a row of $\bhat$.

\begin{table}[h!] 
\centering
\begin{tabular}{c|c|ccc|ccc}
\toprule
\multicolumn{8}{c}{Partial Scenario}\\
\midrule
Method & Truth & Bonf,$\al$=1 & Indep,cpc=0.1 & Bonf-MIC & BH,$\al$=0.3 & BHMat,$\al$=0.3 & BH-MIC \\
\midrule
Train Err & $ 0.31 \pm 0.00 $ & $ 0.52 \pm 0.01 $ & $ 0.63 \pm 0.01 $ & $ 0.56 \pm 0.01 $ & $ 0.51 \pm 0.01 $ & $ 0.52 \pm 0.01 $ & $ 0.53 \pm 0.01 $ \\
Test Err & $ 0.32 \pm 0.00 $ & $ {\bf 0.57 \pm 0.01} $ & $ 0.66 \pm 0.01 $ & $ 0.59 \pm 0.02 $ & $ {\bf 0.56 \pm 0.01} $ & $ 0.57 \pm 0.01 $ & $ 0.57 \pm 0.01 $ \\
Coeff Prec & $ 1.00 \pm 0.00 $ & $ 0.74 \pm 0.01 $ & $ 0.96 \pm 0.00 $ & $ 0.76 \pm 0.01 $ & $ 0.70 \pm 0.01 $ & $ 0.73 \pm 0.01 $ & $ 0.74 \pm 0.01 $ \\
Coeff Rec & $ 1.00 \pm 0.00 $ & $ 0.60 \pm 0.01 $ & $ 0.53 \pm 0.01 $ & $ {\bf 0.71 \pm 0.01} $ & $ 0.61 \pm 0.01 $ & $ 0.60 \pm 0.01 $ & ${\bf 0.73 \pm 0.01} $ \\
\midrule
\multicolumn{8}{c}{Full Scenario}\\
\midrule
Method & Truth & Bonf,$\al$=2 & Indep,cpc=0.1 & Bonf-MIC & BH,$\al$=0.45 & BHMat,$\al$=0.5 & BH-MIC \\
\midrule
Train Err & $ 0.31 \pm 0.00 $ & $ 0.49 \pm 0.01 $ & $ 0.61 \pm 0.01 $ & $ 0.30 \pm 0.00 $ & $ 0.49 \pm 0.01 $ & $ 0.48 \pm 0.01 $ & $ 0.30 \pm 0.00 $ \\
Test Err & $ 0.32 \pm 0.00 $ & $ 0.54 \pm 0.01 $ & $ 0.63 \pm 0.01 $ & $ {\bf 0.33 \pm 0.00} $ & $ 0.54 \pm 0.01 $ & $ 0.53 \pm 0.01 $ & $ {\bf 0.33 \pm 0.00} $ \\
Coeff Prec & $ 1.00 \pm 0.00 $ & $ 0.62 \pm 0.01 $ & $ 0.97 \pm 0.00 $ & $ 0.62 \pm 0.01 $ & $ 0.59 \pm 0.01 $ & $ 0.57 \pm 0.01 $ & $ 0.61 \pm 0.01 $ \\
Coeff Rec & $ 1.00 \pm 0.00 $ & $ 0.61 \pm 0.01 $ & $ 0.53 \pm 0.01 $ & $ {\bf 0.99 \pm 0.00} $ & $ 0.61 \pm 0.01 $ & $ 0.61 \pm 0.01 $ & $ {\bf 0.99 \pm 0.00} $ \\
\midrule
\multicolumn{8}{c}{Independent Scenario}\\
\midrule
Method & Truth & Bonf,$\al$=0.1 & Indep,cpc=0.1 & Bonf-MIC & BH,$\al$=0.04 & BHMat,$\al$=0.05 & BH-MIC \\
\midrule
Train Err & $ 0.31 \pm 0.00 $ & $ 0.59 \pm 0.01 $ & $ 0.59 \pm 0.01 $ & $ 0.82 \pm 0.02 $ & $ 0.59 \pm 0.01 $ & $ 0.59 \pm 0.01 $ & $ 0.70 \pm 0.01 $ \\
Test Err & $ 0.32 \pm 0.00 $ & $ {\bf 0.62 \pm 0.01} $ & ${\bf  0.62 \pm 0.01}$ & $ 0.86 \pm 0.02 $ & $ {\bf 0.62 \pm 0.01} $ & $ {\bf 0.62 \pm 0.01} $ & $ 0.73 \pm 0.02 $ \\
Coeff Prec & $ 1.00 \pm 0.00 $ & $ 0.96 \pm 0.01 $ & $ 0.96 \pm 0.01 $ & $ 0.96 \pm 0.01 $ & $ 0.96 \pm 0.01 $ & $ 0.96 \pm 0.01 $ & $ 0.96 \pm 0.01 $ \\
Coeff Rec & $ 1.00 \pm 0.00 $ & $ 0.53 \pm 0.01 $ & $ {\bf 0.54 \pm 0.01} $ & $ 0.40 \pm 0.01 $ & $ {\bf 0.54 \pm 0.01} $ & $ {\bf 0.54 \pm 0.01} $ & $ 0.47 \pm 0.01 $ \\
\bottomrule
\end{tabular}
\caption{Test-set accuracy, precision, and recall of MIC and other methods on 25 instances of the synthetic data sets generated as described in Section \ref{Synthetic Data}, except with $m=1{,}000$.}
\label{synthetic}
\end{table}

\subsubsection{Simulated Yeast Data}\label{Simulated Yeast Data}

We created a synthetic Yeast Growth data set based on the one described in Section \ref{Yeast Growth}. As will be explained below, talking about the ``true coefficients" of a synthetic model requires that the features be relatively uncorrelated. However, the transcript features of the original data set were highly correlated, so we included only the 526 marker features in the $X$ matrix. As before, we made use of the 20 growth clusters as the $Y$ matrix. We created four types of data sets with varying degrees of correlation among the features, but the basic data-generation process was the same for each, and we describe it below.

We started by trying to approximate what the true distribution of nonzero coefficients in $\beta$ might look like by running the RIC algorithm described in Section \ref{Information-Theoretic Regression---Many Responses} with a cost of 3 bits per coefficient. The resulting $526 \times 20$ $\bhat$ matrix contained 33 nonzero coefficients, many of them in blocks of four or five down certain columns due to correlation among adjacent marker features.

The next step was to construct a true $\beta$ matrix $\beta_\text{sim}$ resembling, but not identical to, $\bhat$.\footnote{If we took $\beta_\text{sim}$ to be equal to $\bhat$, then the RIC algorithm might have artificially high precision and recall. That's because we're defining the ``true features" for the synthetic data set as those that the RIC algorithm returned on the real data set. Since we designed the synthetic data set to imitate the real one, we might expect RIC to return similar patterns of coefficients in both cases. Generating a new matrix to serve as $\beta_\text{sim}$ also allows for randomization over multiple instances of this synthetic data set. Of course, one downside of this simulation approach is that correlational structure from the original problem is lost.} We did this based on summary statistics about $\bhat$: What fraction $f$ of the features had any nonzero coefficients? ($f=0.05$ for the actual $\bhat$.) Of those that did, what was the average number $a$ of nonzero coefficients in each row? ($a=1.3$ here.) And what were the sample mean $\mu_{\widehat{\beta}}$ and standard deviation $\sigma_{\widehat{\beta}}$ of the nonzero coefficient values? (Actual values were $-0.12$ and $0.20$, respectively.) We initialized an empty $526 \times 20$ matrix $\beta_\text{sim}$ and, to fill it in, walked down the rows, each time flipping a coin with probability $f$ to decide whether to give that row nonzero coefficients. We drew the number of nonzero coefficients from a Poisson distribution with rate $a$ (capped at $20$, the total number of responses) and distributed those coefficients randomly among the $20$ columns. The values of each of those coefficients were drawn independently from a normal distribution with mean $\mu_{\widehat{\beta}}$ and standard deviation $\sigma_{\widehat{\beta}}$. When we had finished walking down the rows, we checked to make sure that the total number of nonzero coefficients in $\beta_\text{sim}$ was within 25\% of that in the original $\bhat$; if not, we started the process over.

\begin{figure}[h]
\begin{minipage}{0.29\linewidth}
\centering
\includegraphics[scale=.15]{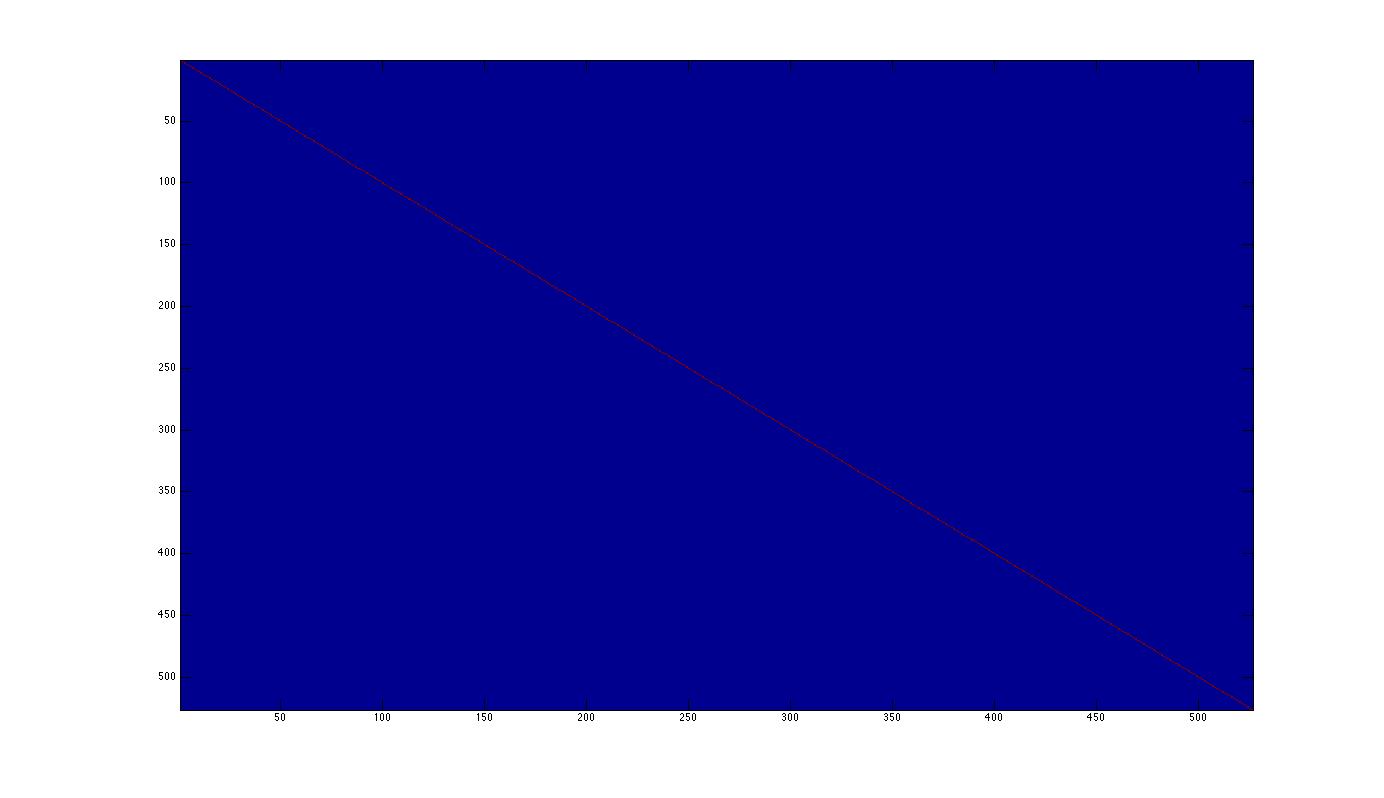}
\label{all}
\end{minipage}
\hspace{0.014\linewidth} 
\begin{minipage}{0.29\linewidth}
\centering
\includegraphics[scale=.15]{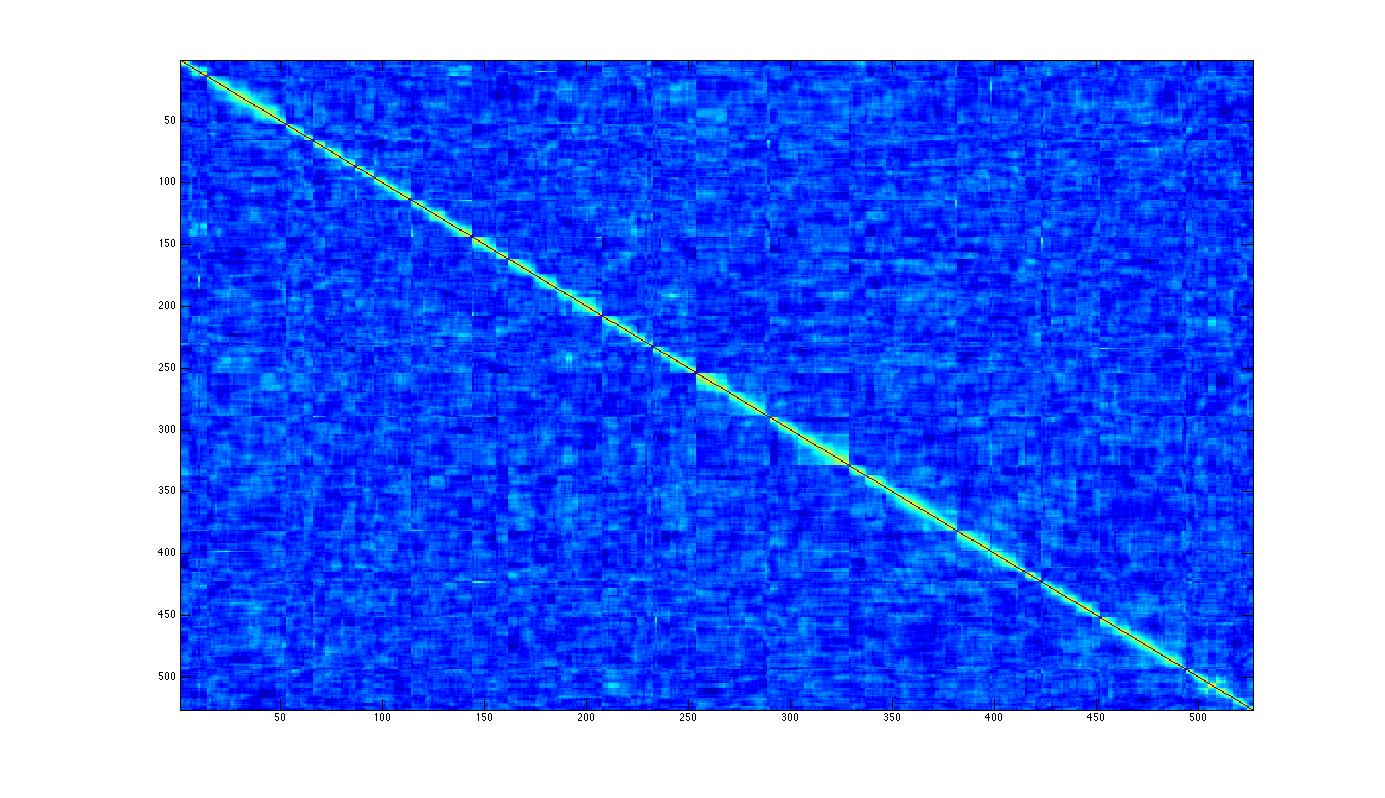}
\label{half}
\end{minipage}
\hspace{0.014\linewidth} 
\begin{minipage}{0.29\linewidth}
\centering
\includegraphics[scale=.15]{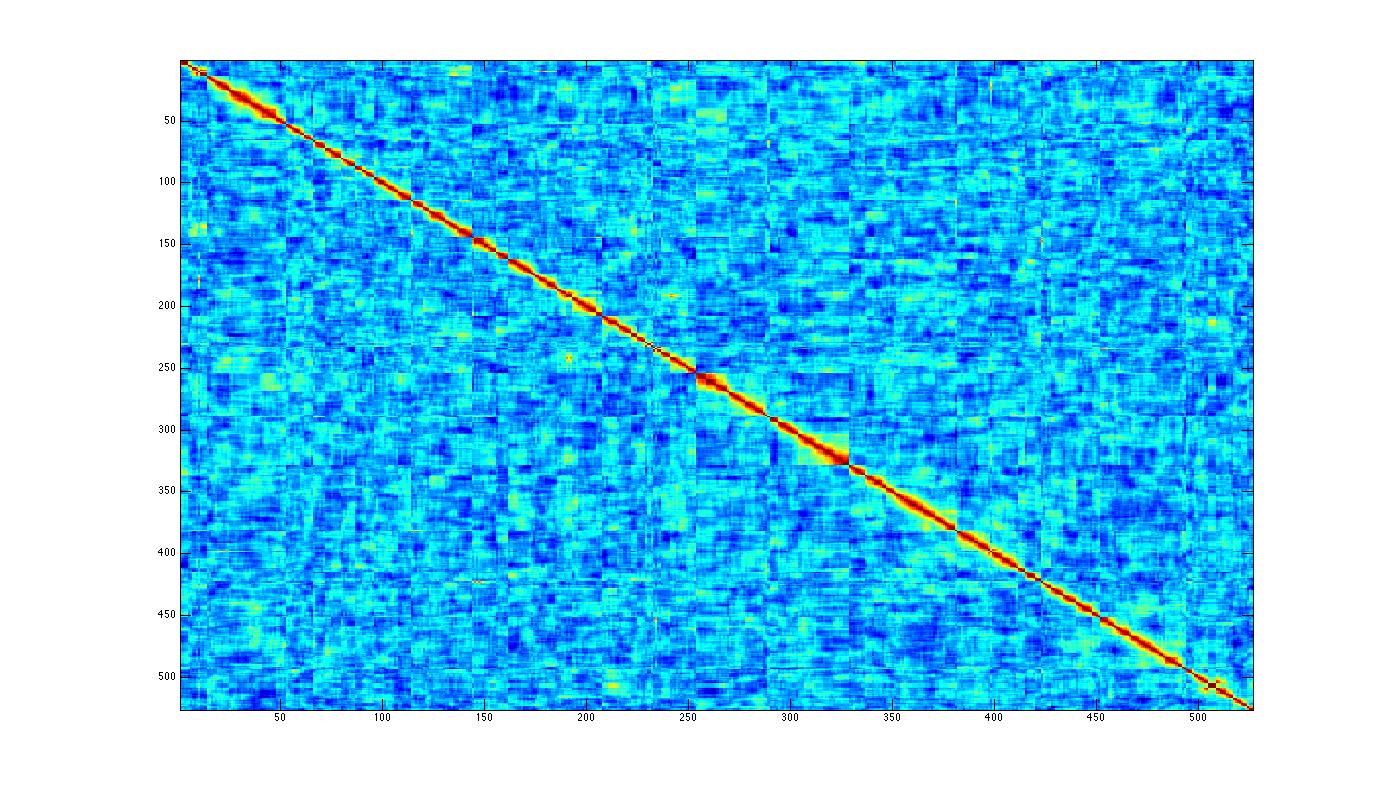}
\label{no}
\end{minipage}
\caption{Correlation matrices of the 526 marker features in the Yeast Growth data set. The figures correspond to, from left to right, diagonal shrinkage, half-diagonal shrinkage, and no shrinkage. Red = correlation near 1, yellow = correlation near 0.8, light blue = correlation near 0.2, dark blue = correlation near 0. Due to the relatively high correlation among features, measures of precision and recall are slightly misleading, because which features are ``truly correlated" with responses becomes unclear, as the text explains.}
\label{corrMatricesYeast}
\end{figure}

We then constructed a simulated $100 \times 526$ $X$ matrix $X_\text{sim}$ by drawing each row from a multivariate-normal distribution with some covariance $\widehat{\Sigma}^X$ based on the covariance matrix of the real $X$ matrix. The first three variants of our synthetic data set consisted of taking $\widehat{\Sigma}^X$ to be one of $\widehat{\Sigma}_D^X$ (diagonal), $\widehat{\Sigma}_\lambda^X$ with $\lambda = 0.5$ (half diagonal, half full), and $\widehat{\Sigma}_F^X$ (full), with the subscripts as in Section \ref{logLikeMultipleY}. Figure \ref{corrMatricesYeast} plots, respectively, $\widehat{\Sigma}_D^X$, $\widehat{\Sigma}_{0.5}^X$, and $\widehat{\Sigma}_F^X$ as correlation matrices (i.e., each entry is scaled to fall in the range $[0,1]$), with red indicating ``high correlation" and blue, ``low correlation."

All three of the above variants involved normally distributed feature values, even though $X$ itself consisted only of 0's and 1's (minor and major allele, respectively). Thus, as the fourth variant, we took $X_\text{sim}$ to be the original binary-valued $X$ matrix.

Finally, we estimated $\widehat{\Sigma}_D$ in the manner of Section \ref{logLikeMultipleY} with only an intercept feature in the model. With these matrices in place, we computed
\s{
Y_\text{sim} = X_\text{sim} \beta_\text{sim} + \eps_\text{sim},
}
with each row of  $\eps_\text{sim}$ being independently distributed as $\mathcal{N}_h ( 0, \widehat{\Sigma}_D )$.

Table \ref{growthbasedsynthetic} shows the results for each variant of the data set. In general, MIC performs essentially indistinguishably from the other methods. This is probably because $a$, the average number of responses per feature, is only 1.3 for this particular data set, implying that there isn't substantial sharing of features across responses. There does appear to be enough sharing that MIC doesn't perform as much worse than the other methods as in the case of the Independent synthetic data set in Table \ref{synthetic}.

The precision of each of the methods degrades significantly as the correlation among the features increases. In fact, these numbers are somewhat misleading, because the notion of a ``true coefficient" becomes fuzzy when features are correlated. We generated the data by imposing a particular $\beta_\text{sim}$ matrix, with nonzero coefficients in certain locations. For instance, for response 3, we might have given feature 186 a coefficient of 0.2, causing a correlation between that feature and that response. However, because feature 186 is correlated with, say, features 185 and 187, features 185 and 187 will also be correlated with response 3, so we end up selecting them as well. This isn't necessarily wrong to do, because the data really do show a correlation; the arbitrariness of our choice of zeros and nonzeros in $\beta_\text{sim}$ is at fault for the large apparent rate of ``false positives."

\begin{table}[h!] 
\centering
\begin{tabular}{c|c|ccc|ccc}
\toprule
\multicolumn{8}{c}{Each row of $X$ distributed $\mathcal{N}_m ( 0, \widehat{\Sigma}_D^X )$---no correlation among features. }\\
\midrule
Method & Truth & Bonf,$\al$=0.1 & Indep,cpc=0.1 & Bonf-MIC & BH,$\al$=0.06 & BHMat,$\al$=0.085 & BH-MIC \\
\midrule
Train Err & $ 0.02 \pm 0.00 $ & $ 0.02 \pm 0.00 $ & $ 0.02 \pm 0.00 $ & $ 0.02 \pm 0.00 $ & $ 0.02 \pm 0.00 $ & $ 0.02 \pm 0.00 $ & $ 0.02 \pm 0.00 $ \\
Test Err & $ 0.02 \pm 0.00 $ & $ 0.03 \pm 0.00 $ & $ {\bf 0.02 \pm 0.00} $ & $ 0.03 \pm 0.00 $ & $ {\bf 0.02 \pm 0.00} $ & $ {\bf 0.02 \pm 0.00} $ & $ {\bf 0.02 \pm 0.00} $ \\
Coeff Prec & $ 1.00 \pm 0.00 $ & $ 0.92 \pm 0.01 $ & $ 0.90 \pm 0.01 $ & $ 0.92 \pm 0.01 $ & $ 0.91 \pm 0.01 $ & $ 0.91 \pm 0.01 $ & $ 0.92 \pm 0.01 $ \\
Coeff Rec & $ 1.00 \pm 0.00 $ & $ 0.80 \pm 0.01 $ & $ 0.81 \pm 0.01 $ & $ 0.79 \pm 0.01 $ & $ {\bf 0.81 \pm 0.01} $ & $ {\bf 0.81 \pm 0.01} $ & $ 0.80 \pm 0.01 $ \\
\midrule
\multicolumn{8}{c}{Each row of $X$ distributed $\mathcal{N}_m ( 0, \widehat{\Sigma}_{0.5}^X )$---some correlation among features. }\\
\midrule
Method & Truth & Bonf,$\al$=0.055 & Indep,cpc=1 & Bonf-MIC & BH,$\al$=0.03 & BHMat,$\al$=0.045 & BH-MIC \\
\midrule
Train Err & $ 0.02 \pm 0.00 $ & $ 0.03 \pm 0.00 $ & $ 0.03 \pm 0.00 $ & $ 0.02 \pm 0.00 $ & $ 0.02 \pm 0.00 $ & $ 0.02 \pm 0.00 $ & $ 0.02 \pm 0.00 $ \\
Test Err & $ 0.02 \pm 0.00 $ & $ {\bf 0.03 \pm 0.00} $ & $ {\bf 0.03 \pm 0.00} $ & $ {\bf 0.03 \pm 0.00} $ & $ 0.03 \pm 0.00 $ & $ 0.03 \pm 0.00 $ & $ {\bf 0.02 \pm 0.00} $ \\
Coeff Prec & $ 1.00 \pm 0.00 $ & $ 0.78 \pm 0.01 $ & $ 0.76 \pm 0.01 $ & $ 0.78 \pm 0.01 $ & $ 0.75 \pm 0.01 $ & $ 0.75 \pm 0.01 $ & $ 0.75 \pm 0.01 $ \\
Coeff Rec & $ 1.00 \pm 0.00 $ & $ 0.79 \pm 0.01 $ & $ 0.79 \pm 0.01 $ & $ {\bf 0.80 \pm 0.01} $ & $ 0.79 \pm 0.01 $ & $ 0.79 \pm 0.01 $ & $ {\bf 0.80 \pm 0.01} $ \\
\midrule
\multicolumn{8}{c}{Each row of $X$ distributed $\mathcal{N}_m ( 0, \widehat{\Sigma}_F^X )$---lots of correlation among features. }\\
\midrule
Method & Truth & Bonf,$\al$=0.19 & Indep,cpc=0.1 & Bonf-MIC & BH,$\al$=0.07 & BHMat,$\al$=0.06 & BH-MIC \\
\midrule
Train Err & $ 0.02 \pm 0.00 $ & $ 0.02 \pm 0.00 $ & $ 0.02 \pm 0.00 $ & $ 0.02 \pm 0.00 $ & $ 0.02 \pm 0.00 $ & $ 0.02 \pm 0.00 $ & $ 0.02 \pm 0.00 $ \\
Test Err & $ 0.02 \pm 0.00 $ & $ {\bf 0.02 \pm 0.00} $ & $ {\bf 0.02 \pm 0.00} $ & $ {\bf 0.02 \pm 0.00} $ & $ {\bf 0.02 \pm 0.00} $ & $ {\bf 0.02 \pm 0.00} $ & $ {\bf 0.02 \pm 0.00} $ \\
Coeff Prec & $ 1.00 \pm 0.00 $ & $ 0.19 \pm 0.01 $ & $ 0.20 \pm 0.01 $ & $ 0.19 \pm 0.01 $ & $ 0.15 \pm 0.01 $ & $ 0.15 \pm 0.00 $ & $ 0.15 \pm 0.01 $ \\
Coeff Rec & $ 1.00 \pm 0.00 $ & $ {\bf 0.84 \pm 0.01} $ & $ {\bf 0.84 \pm 0.01} $ & $ {\bf 0.84 \pm 0.01} $ & $ {\bf 0.86 \pm 0.01} $ & $ 0.85 \pm 0.01 $ & $ 0.85 \pm 0.01 $ \\
\midrule
\multicolumn{8}{c}{Real, original $X$ matrix. }\\
\midrule
Method & Truth & Bonf,$\al$=0.17 & Indep,cpc=0.1 & Bonf-MIC & BH,$\al$=0.06 & BHMat,$\al$=0.055 & BH-MIC \\
\midrule
Train Err & $ 0.02 \pm 0.00 $ & $ 0.02 \pm 0.00 $ & $ 0.02 \pm 0.00 $ & $ 0.02 \pm 0.00 $ & $ 0.02 \pm 0.00 $ & $ 0.02 \pm 0.00 $ & $ 0.02 \pm 0.00 $ \\
Test Err & $ 0.02 \pm 0.00 $ & $ {\bf 0.02 \pm 0.00}  $ & $ {\bf 0.02 \pm 0.00} $ & $ {\bf 0.02 \pm 0.00}  $ & $ {\bf 0.02 \pm 0.00}  $ & $ {\bf 0.02 \pm 0.00}  $ & $ {\bf 0.02 \pm 0.00}  $ \\
Coeff Prec & $ 1.00 \pm 0.00 $ & $ 0.15 \pm 0.00 $ & $ 0.15 \pm 0.00 $ & $ 0.15 \pm 0.00 $ & $ 0.13 \pm 0.00 $ & $ 0.13 \pm 0.00 $ & $ 0.13 \pm 0.00 $ \\
Coeff Rec & $ 1.00 \pm 0.00 $ & $ 0.82 \pm 0.01 $ & $ 0.81 \pm 0.01 $ & $ {\bf 0.83 \pm 0.01 } $ & $ {\bf 0.84 \pm 0.01} $ & $ {\bf 0.84 \pm 0.01} $ & $ {\bf 0.84 \pm 0.01} $ \\
\bottomrule
\end{tabular}
\caption{ Test-set accuracy, precision, and recall of MIC and other methods on 25 instances of the synthetic data sets generated as described in Section \ref{Simulated Yeast Data}, with $m=526$ features.}
\label{growthbasedsynthetic}
\end{table}

\section{Conclusion}\label{conclusion}

The MDL principle provides a natural framework in which to design  penalized-regression criteria for feature selection. In the case of a single response, one example of this is RIC, which can be characterized by an information-theoretic penalty of $2 \lg m$ bits per feature when selecting from among $m$ total features. We proposed an extension of this criterion, called MIC, for the case of multiple responses. By efficiently coding the locations of feature-response pairs for features associated with multiple responses, MIC allows for sharing of information across responses during feature selection. The method is competitive with, and sometimes outperforms, existing multitask learning algorithms in terms of prediction accuracy, while achieving generally sparser, more interpretable models.

MDL can also be viewed in the domain of hypothesis testing as a way of correcting against alpha inflation in multiple tests. We explained how MDL regression applied to each feature separately can be interpreted along the lines of standard Bonferroni and Benjamini-Hochberg feature-selection procedures. Again using the MIC approach, we extended this to hypothesis testing with multiple responses, allowing for greater power in selecting true coefficients when features are significantly shared across responses.


\appendix
\section{MDL and Hypothesis Testing}

Information theory describes an isomorphism between probabilities and code lengths in which the idealized code length of some symbol $x$ is $- \lg P(x)$  \cite[p. 28]{grunwald2005}. There is a similar, though only approximate, relationship between MDL and the process of \textit{statistical hypothesis testing}: namely, that MDL will tend to introduce an extra parameter into a model in roughly the same cases that a hypothesis test would reject the null hypothesis that the parameter is zero. This is perhaps unsurprising, because MDL is a tool for model selection, and hypothesis testing is about rejecting models that fit the data poorly.

\subsection{Generalized Likelihood Ratio Tests}

Given two models $\model_0$ and $\model_1$ and data $\data$, statisticians define $\Lambda$ to be the ratio of their likelihoods:
\sn{def-Lambda}{
\Lambda := \f{P(\data \giv \model_0)}{P(\data \giv \model_1)}.
}
In many cases, these models correspond to the same probability density function with different parameter settings. For instance, in regression of a single response $y$ on a single feature $x$ (with no intercept for simplicity):
\sn{simpleRegEx}{
y = \beta x + \eps, \ \ \eps \sim \mathcal{N}(0, \sigma^2),
}
$\model_0$ might specify that, given $x$, $y$ is distributed $\mathcal{N}(0, \sigma^2)$, i.e., that $\beta = 0$, while $\model_1$ might say that given $x$, $y$ is distributed $\mathcal{N}(\bhat x, \sigma^2)$, where $\bhat$ is the maximum-likelihood estimate for $\beta$ derived from the training data. Note that $\model_1$ here depends on the observed data $\data$. Some authors prefer to speak of fixed probability distributions $\model_\theta$ with parameters $\theta$ ranging over different parameter spaces $\Theta_0$ and $\Theta_1$ \cite[p. 463]{larsen_introduction_2005}, in which case
\s{
\Lambda = \f{\displaystyle \sup_{\theta \in \Theta_0} P(\data \giv \model_\theta)}{\displaystyle \sup_{\theta \in \Theta_1} P(\data \giv \model_\theta)}.
}
In the regression example, $\Theta_0 = \set{0}$ while $\Theta_1 = \mathbb{R}$, the entire real line.

In fact, it will be convenient to consider $-\lg \Lambda$, which is large when $\Lambda$ is small. In some sense,  $-\lg \Lambda$ measures the ``badness of fit" of $\model_0$ relative to $\model_1$. Over different data sets $\data$,  $-\lg \Lambda$ will take on different values, according to some probability distribution. If $\model_0$ is true,  $-\ln \Lambda$ is unlikely to be very large, so we can define a threshold $T_\al$ such that
\sn{talpha}{
P(- \lg \Lambda >  T_\alpha \giv \model_0) = \alpha
}
and reject $\model_0$ whenever $-\ln \Lambda$ exceeds this threshold. This is known as a \textit{generalized likelihood ratio test} (GLRT), of which many standard statistical hypothesis tests are examples, including the $t$-test on the significance of a regression coefficient \cite[p. 685]{larsen_introduction_2005}.

\subsection{Model Selection}

Consider the approach that MDL would take. It ``rejects" $\model_0$ in favor of $\model_1$ just in the case that the description length associated with $\model_1$ is shorter:
\sn{choose1}{
\ell(\model_1) + \ell(\data \giv \model_1) < \ell(\model_0) + \ell(\data \giv \model_0),
}
where $\ell(\cdot)$ stands for description length. By the discussion in section \ref{Coding the Data}, $\ell(\data \giv \model_1) = - \lg P(\data \giv \model_1)$, so that \eqref{choose1} becomes
\s{
\ell(\model_1) - \ell(\model_0)  <  \lg P(\data \giv \model_1) - \lg P(\data \giv \model_0),
}
or
\sn{choose1again}{
- \lg \Lambda > \ell(\model_1) - \ell(\model_0).
}
If $\model_1$ is more complicated than $\model_0$, then $\ell(\model_1) - \ell(\model_0)$ will be positive, and there will be some $\al$ at which $\ell(\model_1) - \ell(\model_0) = T_\al$ as defined in \eqref{talpha}. Thus, selecting between $\model_0$ and $\model_1$ is equivalent to doing a likelihood ratio test at the implicit significance level $\al$ determined by $\ell(\model_1) - \ell(\model_0)$.

\subsection{Example: Single Regression Coefficient}\label{Example: Single Regression Coefficient}

While MDL introduces the quantity $- \lg \Lambda$, it is more common in statistics to deal instead with $-2 \ln \Lambda = -(2 \ln 2) \lg \Lambda$ because of the following result: If $\model_0$ and $\model_1$ both belong to the same type of probability distribution satisfying certain smoothness conditions, and if $\model_0$ has parameter space $\Theta_0$ with dimensionality $\text{dim}(\Theta_0)$ while $\model_1$ has parameter space $\Theta_1$ with dimensionality $\dim(\Theta_1) > \dim(\Theta_0)$, then under $\model_1$, $-(2 \ln 2)  \lg \Lambda$ is asymptotically chi-square with degrees of freedom equal to $\dim(\Theta_1) - \dim(\Theta_0)$ \cite[sec. 9.5]{rice1995msa}. 
In particular, if the probability distribution is normal (as in our regression model), then the chi-square distribution will be not just asymptotic but exact. 

In \eqref{simpleRegEx}, where we have a single regression coefficient, $\dim(\Theta_1) - \dim(\Theta_0) = 1$, so that under $\model_1$, $-(2 \ln 2)  \lg \Lambda$ has a chi-square distribution with 1 degree of freedom, which is the same as the distribution of the square of a standard normal random variable $Z \sim \mathcal{N}(0, 1)$. Thus, we can rewrite \eqref{talpha} as
\sn{talpha2}{
P\paren{- (2 \ln 2) \lg \Lambda >  (2 \ln 2) T_\alpha \giv \model_0} = \alpha \ & \Longleftrightarrow \ P(Z^2 > (2 \ln 2) T_\alpha) = \alpha\\
\ & \Longleftrightarrow \ P(Z < -\sqrt{(2 \ln 2) T_\alpha}) = \f{\alpha}{2}\\
& \Longleftrightarrow \ \Phi(-\sqrt{(2 \ln 2) T_\alpha}) = \f{\alpha}{2},
}
where $\Phi$ is the cumulative distribution function of the standard-normal distribution. Using the approximation $\Phi(-x) \approx \frac{1}{4} e^{-\f{x^2}{2}}$ for relatively large $x > 0$,\footnote{This follows from approximating
\s{
\Phi(-x) = \int_{-\infty}^{-x} \f{1}{\sqrt{2 \pi}} e^{-\f{t^2}{2}} \, \dd t
}
by the value of the integrand at $t=-x$, assuming $\f{1}{\sqrt{2 \pi}} \approx \f{1}{4}$.

Another way to see the approximation is as follows. \cite[pp. 63-67]{polya1945rcp} proved that for $x \geq 0$,
\sn{polya}{
0.5 - \Phi(-x) \approx \half \sqrt{1 - \exp\paren{-\frac{2 x^2}{\pi}}}.
}
If $x$ is large enough that $\Phi(-x)^2$ is negligible compared with $\Phi(-x)$, then we can multiply by 2 and square both sides of \eqref{polya} to give
\s{
1 - 4 \Phi(-x) \approx 1 - \exp\paren{-\frac{2 x^2}{\pi}}  \ \Longleftrightarrow \ \Phi(-x) \approx \f{1}{4} \exp\paren{-\frac{2 x^2}{\pi}}
}
Assuming $\pi \approx 4$ gives the result.

Note that $\Phi(-x) \approx \frac{1}{x \sqrt{2 \pi}} e^{-\f{x^2}{2}}$ tends to give a tighter approximation, especially for $x$ bigger than $\sim 2$, but it's less mathematically convenient here.

 I credit \cite[p. 17]{stine-tutorial} with inspiring the approach of using an approximation to $\Phi$ to compare hypothesis testing and coding costs.}  \eqref{talpha2} becomes
\sn{final}{
\frac{1}{4} \exp\paren{-\f{(2 \ln 2) T_\alpha}{2}} = \f{\al}{2} \ \Longleftrightarrow \ T_\al = - \f{ \ln (2 \al) }{\ln 2} = - \lg \al - 1.
}

\subsection{Bonferroni Criterion}\label{bonferroni-appendix}
The Bonferroni criterion in p-value space says ``reject $H_0$ when the p-value is less than $\frac{\alpha}{m}$." By \eqref{talpha}, we can express this in log-likelihood space by saying ``reject $H_0$ when $- \lg \Lambda > T_\f{\al}{m}$," where, by \eqref{final},
\sn{bonferroni-lgm}{
T_\f{\al}{m} =  \lg m - \lg \al - 1.
}
Note the similarity to the decision criterion \eqref{bonf-decision}.

\subsection{Closeness of the Approximation}\label{closeness-approx}

It's worth reflecting on why these correspondences are only approximate. Indeed, if code length is just negative log probability, and a p-value $p$ is a probability, then, say, the Bonferroni rule of rejecting when $p < \f{\al}{m}$ is equivalent to rejecting when $- \lg p > \lg m - \lg \al$, which looks basically the same as \eqref{bonferroni-lgm}. However, $\Lambda$, which is used by MDL, is not exactly equal to $p$; the former is a comparison of a probability density function at two points, while the latter is an area under the probability density function in extreme regions. Still, the two are often quite close in practice.

\begin{figure}[h]
\begin{center}
\includegraphics[scale=0.4]{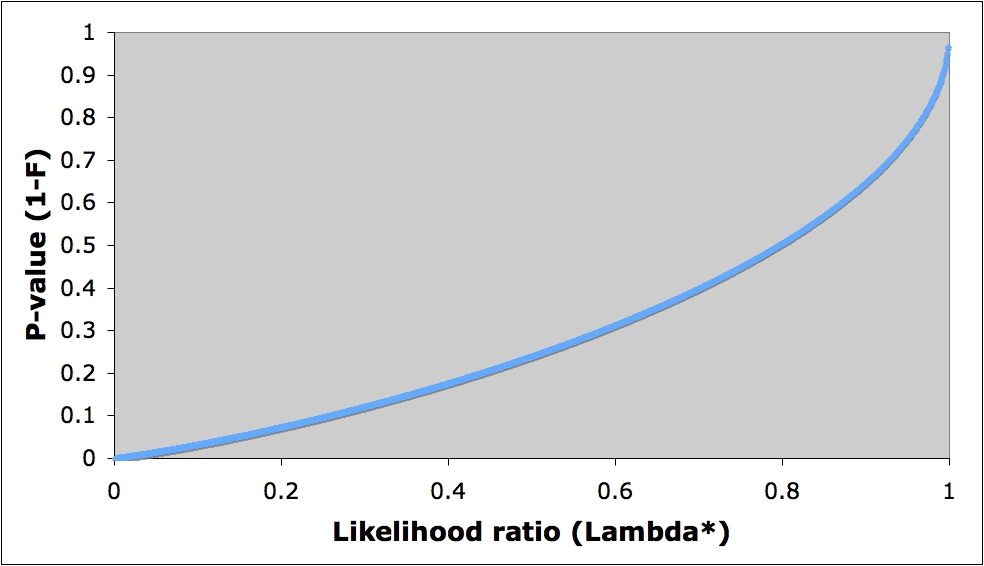}
\end{center}
\caption{The actual p-value $P(\Lambda < \Lambda_* \giv \model_0)$ vs. $\Lambda_*$ for $k=1$ degree of freedom.}
\label{closeness-approx.jpg}
\end{figure}

How close? Suppose we observe some value $\Lambda_*$ for $\Lambda$. The associated p-value is
\s{
p = P(\Lambda < \Lambda_* \giv \model_0) = P(- 2 \ln \Lambda > -2 \ln \Lambda_* \giv \model_0).
}
If $-2 \ln \Lambda \sim \chi^2_{(k)}$ under $\model_0$,
\sn{pOfLambda}{
p = 1-F_{\chi^2_{(k)}} \paren{ -2 \ln \Lambda_* },
}
where $F$ stands for the cumulative distribution function. Figure \ref{closeness-approx.jpg} compares $\Lambda_*$ to the associated p-value for the case of $k=1$ degree of freedom; a straight, 45-degree line would be a perfect match, but the approximation is generally correct to within a factor of 2 or 3.

We can use this curve to compute the $\al$ implied by a more complicated model $\model_1$ in \eqref{choose1again}. Letting $\Delta c := \ell(\model_1) - \ell(\model_0)$, \eqref{choose1again} and \eqref{pOfLambda} give
\s{
\al = 1-F_{\chi^2_{(k)}} \paren{ (2 \ln 2) \Delta c }.
}
Table \ref{exAlphas} shows examples for $k=1$ degree of freedom. $\al=0.05$ is achieved at $\Delta c = 2.77$ bits.\footnote{\cite[p. 582]{chapter-maimon2005dma} present a very similar discussion, only with log base $e$ and with their $\lambda$ equal to $\frac{\Delta c}{2}$. The numerical values given are equivalent. }

\begin{table}[h]
\centering
\begin{tabular}{ccccc}
\toprule 
$\Delta c$ (bits) & 1 & 2 & 3 & 4\\
Implied $\al$ & 0.24 & 0.1 & 0.04 & 0.02\\
\bottomrule
\end{tabular}
\caption{Example implied $\al$ values against the increase in coding cost $\Delta c$ of the more complicated model, for $k=1$ degree of freedom.}
\label{exAlphas}
\end{table}


\subsection{BH-Style Penalty}\label{BH-appendix}

Suppose we regress $y$ on each of $m$ features separately. We would obtain $m$ p-values, to which we could apply the BH step-up procedure \eqref{BH-procedure}. Alternatively, we could evaluate, for each feature $j$, a negative log-likelihood ratio:
\s{
-\lg \Lambda_j = - \lg P(Y \giv \model_0) - \paren{ - \lg P(Y \giv \model_{1j}) },
}
where $\model_{1j}$ denotes the alternative hypothesis in which feature $j$ has a nonzero, maximum-likelihood coefficient.

In the same way that we obtained \eqref{bonferroni-lgm}, we can derive a BH-type rejection level of $\frac{j \alpha}{m}$:
\s{
T_\f{j \al}{m} =  \lg m - \lg j - \lg \al - 1.
}
We can now restate the BH procedure as follows: Put the $- \lg \Lambda_j$ in decreasing order, and let $- \lg \Lambda_{(j)}$ be the $j^\text{th}$ biggest. Reject $H_{(1)}, \ldots, H_{(q)}$ such that
\sn{BHwithlambda}{
q = \max \set{j : - \lg \Lambda_{(j)} \geq \lg m - \lg j - \lg \al - 1}.
}
We could rephrase \eqref{BHwithlambda} as choosing the $q$ that maximizes
\s{
\sum_{j=1}^q \paren{ - \lg \Lambda_{(j)} - \lg m + \lg j + \lg \alpha + 1 }
}
or that minimizes
\sn{BHwithlambda2}{
-\sum_{j=1}^q \paren{ - \lg \Lambda_{(j)} - \lg m + \lg j + \lg \alpha + 1 }
 = \lg \frac{m^q}{q!} - q (\lg \alpha + 1) + \sum_{j=1}^q \paren{ \lg P(Y \giv \model_0)  - \lg P(Y \giv \model_{1(j)}) }.
}
Since $\lg P(Y \giv \model_0)$ is a constant with respect to $q$, we can subtract it $m$ times from \eqref{BHwithlambda2} to give
\sn{BHwithlambda3}{
\lg \frac{m^q}{q!} - q (\lg \alpha + 1) + \sum_{j=1}^q - \lg P(Y \giv \model_{1(j)})  + \sum_{j=q+1}^m - \lg P(Y \giv \model_0) = \lg \frac{m^q}{q!} - q (\lg \alpha + 1) + \mathcal{D}^m(Y \giv \bhat_q)
}
with $\mathcal{D}^m(Y \giv \bhat_q)$ as in \eqref{dm}, in which $\bhat_q$ is the model containing the $q$ features with highest negative log-likelihood ratio values. Now, if $m \gg q$, then $m^q \approx \frac{m!}{(m-q)!}$, and \eqref{BHwithlambda3} becomes
\sn{BHwithlambda4}{
\lg \frac{m!}{q!(m-q)!} - q (\lg \alpha + 1) + \mathcal{D}^m(Y \giv \bhat_q) = \lg \binom{m}{q} - q (\lg \alpha + 1) + \mathcal{D}^m(Y \giv \bhat_q),
}
in very close analogy to \eqref{BH-mdl-eqn}.\footnote{The point that MDL can reproduce BH-style step-up procedures has been noted elsewhere (e.g., \cite[p. 19]{stine-tutorial}) in slightly different words.}

\section*{Acknowledgements}

The MIC approach to regression for prediction (i.e., the content of Section \ref{regression-section}) was originated by Jing Zhou, Dean Foster, and Lyle Ungar, who drafted a paper outlining the theory and some preliminary experiments. Lyle suggested that I continue this work as a project for summer 2008, focusing on hypothesis testing (Section \ref{hypothesis-mic}). With continued guidance from Lyle and Dean, I kept working on the research into the fall and spring as part of this thesis. In January 2009, Paramveer Dhillon, Lyle, and I submitted a paper to the International Conference on Machine Learning (ICML 2009) highlighting the experimental results in Section \ref{regression-section}. The results in Section \ref{hypothesis-mic} are original to this thesis.

In addition to the names above, I thank Robert Stine and Phil Everson for advice on statistical theory; Qiao Liu for conversations on MIC with Lyle; Dana Pe'er and her laboratory for providing the Yeast data set; Adam Ertel and Ted Sandler for making accessible the Breast Cancer data set;  Jeff Knerr for computing assistance; Tia Newhall, Rich Wicentowski, and Doug Turnbull for guidance on writing this thesis; and Santosh S. Venkatesh, my Honors thesis examiner, for several helpful comments and corrections.

\bibliographystyle{alpha}
\bibliography{thesis}
\end{document}